%% file: main.tex
\documentclass[journal]{IEEEtran}
\ifCLASSINFOpdf
\else
\fi

\usepackage{amsmath,amsfonts}
\usepackage{algorithmic}
\usepackage{algorithm}
\usepackage{array}
\usepackage[caption=false,font=normalsize,labelfont=sf,textfont=sf]{subfig}
\usepackage{textcomp}
\usepackage{stfloats}
\usepackage{url}
\usepackage{verbatim}
\usepackage{graphicx}
\usepackage{cite}
\usepackage{multirow}
\usepackage{pdflscape}
\usepackage{siunitx}

\usepackage[table,xcdraw]{xcolor}
\input{ORCIDheader.tex}

\begin{document}
%
\title{Teaching Autonomous Systems Hands-On: \\Leveraging Modular Small-Scale Hardware \\ in the Robotics Classroom}
%
%
%

\author{Johannes Betz*~\orcidicon{0000-0001-9197-2849}, Hongrui Zheng*~\orcidicon{0000-0001-5597-3182}\\
Zirui Zang, Florian Sauerbeck, Y. Rosa Zheng~\orcidicon{0000-0001-5237-1485}\, Joydeep Biswas~\orcidicon{0000-0002-1211-1731},\\Krzysztof Walas~\orcidicon{0000-0002-2800-2716}, Velin Dimitrov, Madhur Behl~\orcidicon{0000-0002-5921-0331}, Venkat Krovi~\orcidicon{0000-0003-2539-896X}, Rahul Mangharam~\orcidicon{0000-0002-3388-8283}
\thanks{Manuscript received XX 2022; revised XX XX, 2022.
Corresponding author: Johannes Betz (email: joebetz@seas.upenn.edu)}
\thanks{This work was sponsored by the  Department of Transportation and the National Science Foundation (NSF) of USA}
\thanks{*: Equal Contribution}
\thanks{J. Betz, H. Zheng, Z. Zang, and R. Mangharam are with the Department of Electrical and Systems Engineering, University of Pennsylvania, Philadelphia, PA, USA (e-mail: joebetz, hongruiz, zzang, rahulm@seas.upenn.edu) }
\thanks{F. Sauerbeck is with the Institute of Automotive Technology, Technical University of Munich, Germany (e-mail: florian.sauerbeck@tum.de)}
\thanks{R. Zheng is with the College of Engineering and Applied Science, Lehigh University, Bethlehem, PA 18015, USA (e-mail: yrz218@lehigh.edu)}
\thanks{J. Biswas is with the Department of Computer Science, University of Texas Austin,  Austin, TX, USA (e-mail: joydeepb@cs.utexas.edu)}
\thanks{K. Walas is with the Institute of Robotics and Machine Intelligence, Poznan University of Technology,  Poland (e-mail: krzysztof.walas@put.poznan.pl)}
\thanks{M. Behl is with the Department of Computer Science, University of Virginia, Charlottesville, VA, USA (e-mail:madhur.behl@virginia.edu)}
\thanks{V. Krovi is with the College of Engineering, Computing and Applied Sciences, Clemson University, Clemson, SC 29634, USA (e-mail:vkrovi@clemson.edu)}
\thanks{V. Dimitrov is with the Toyota Research Institute, Cambridge, MA 02139, USA (e-mail: velin.dimitrov@tri.global)}
}

%
%

\markboth{IEEE Transactions on Learning Technologies,~Vol.~XX, No.~X, XXXXXX~20XX}%
{Shell \MakeLowercase{\textit{et al.}}: Bare Demo of IEEEtran.cls for IEEE Journals}
%



\maketitle

\begin{abstract}
Although robotics courses are well established in higher education, the courses often focus on theory and sometimes lack the systematic coverage of the techniques involved in developing, deploying, and applying software to real hardware. Additionally, most hardware platforms for robotics teaching are low-level toys aimed at younger students at middle-school levels. To address this gap, an autonomous vehicle hardware platform, called F1TENTH, is developed for teaching autonomous systems hands-on. This article describes the teaching modules and software stack for teaching at various educational levels with the theme of ``racing" and competitions that replace exams. The F1TENTH vehicles offer a modular hardware platform and its related software for teaching the fundamentals of autonomous driving algorithms. From basic reactive methods to advanced planning algorithms, the teaching modules enhance students' computational thinking through autonomous driving with the F1TENTH vehicle. The F1TENTH car fills the gap between research platforms and low-end toy cars and offers hands-on experience in learning the topics in autonomous systems. Four universities have adopted the teaching modules for their semester-long undergraduate and graduate courses for multiple years. Student feedback is used to analyze the effectiveness of the F1TENTH platform. More than 80\% of the students strongly agree that the hardware platform and modules greatly motivate their learning, and more than 70\% of the students strongly agree that the hardware enhanced their understanding of the subjects. The survey results show that more than 80\% of the students strongly agree that the competitions motivate them for the course.

\end{abstract}


%
\IEEEpeerreviewmaketitle

\section{Introduction}
\subsection{Autonomous Driving}
Autonomous driving can potentially disrupt our transportation systems as we know them. It is expected that autonomous vehicles will lead to better capacity utilization on our streets, leading to a more effective traffic flow~\cite{Pribyl2019}. Studies show that road safety has increased significantly due to the increased use of assistance systems \cite{Winkle2016}. With fully self-driving vehicles, safety will increase even more since the human, who is a massive factor in causing crashes, is entirely out of the loop. Furthermore, autonomous vehicles could create \$488 billion~\cite{hayes_2020} in annual savings by reducing traffic accidents and additional savings due to reduced fuel costs and therefore reduced emissions~\cite{Wadud2016}. To achieve autonomous capabilities, the senses and actions of a human driver are emulated by suitable sensors, actuators, and respective software. This software consists of modules for perceiving the environment (\textit{perception}), planning a safe path (\textit{planning}), and finally controlling steering and acceleration to follow the calculated safe path (\textit{control})~\cite{pendlton2017}.

\subsection{Hypothesis and Contributions}

\begin{figure*}[h]
\begin{center}
\includegraphics[scale=0.95]{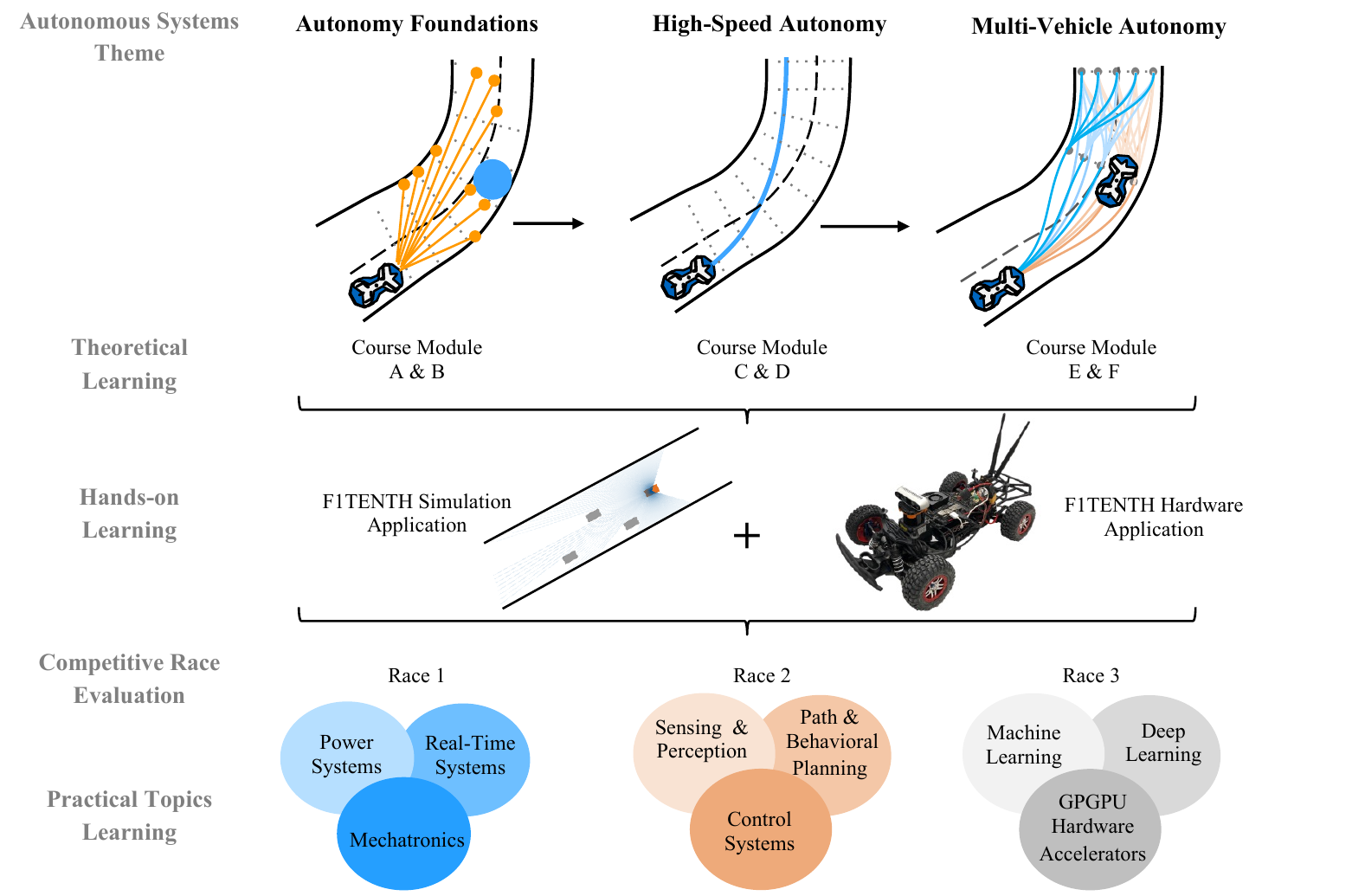}
\caption{The F1TENTH course structure: Teaching autonomous systems hands-on by providing theoretical and practical learning units with simulations and real hardware.}
\label{fig_course_setup}
\end{center}
\end{figure*}
Unfortunately, developing solutions for autonomous driving invokes complexities since it requires well-trained engineers with broad and expert knowledge in machine learning for embedded systems, control theory, and optimization. There will be an increasing demand for specialized engineers since applications for autonomous systems will expand in the future, implying that attracting new students is essential. Therefore, it is crucial to provide relevant and attractive lectures to young students \cite{Renz2021}. Teaching autonomous systems topics at higher education institutions can be seen as a global strategic initiative to educate the next generation of robotics and autonomous systems engineers~\cite{Tang2018}~\cite{Shibata2021}. However, current robotics and autonomous systems courses curricula lack hands-on teaching and actual hardware usage. Literature reviews agree that teaching autonomous systems in higher education that facilitates at an early level needs to be enhanced~\cite{Tang2018}~\cite{Shibata2021}. While the foundations of autonomous systems theory still need to be taught, providing a deeper insight into the software's application on real hardware is essential. Only with this is it possible to educate the next generation of autonomous systems engineers.

To overcome this issue, a new course for teaching hands-on autonomous systems with modular autonomous vehicle hardware and software has been created. Figure \ref{fig_course_setup} displays the content and ideas of this course for teaching autonomous systems in a more applied way. We provide three autonomous systems themes (Foundations; High-Speed; Multi-Vehicle) split into six different course modules (A-F). Here, the students learn the theoretical foundations of autonomous systems. Additionally, a hands-on learning part is provided where the theory of each module is applied in simulation and on the real vehicle hardware. Each theoretical and hands-on learning is tested and evaluated in a race with the actual vehicles. Ultimately, this new course setup provides learnings in various practical topics like mechatronics, control systems, and artificial intelligence.

The hypothesis is that autonomous driving fundamentals must be taught in combination with actual hardware to prepare the students for industry and academia jobs. This combination will enhance the students' \textit{computational thinking} regarding the software and their \textit{systems thinking} regarding the whole autonomous vehicle. This is because the students are allowed for repeated testing and iteration and have the affordance of a physical device to learn as opposed to on-screen simulation only \cite{Papert1980}.
Furthermore, it is hypothesized that by teaching autonomous driving in a competitive environment called \textit{Autonomous Racing}~\cite{Betz2022}, the motivation and fascination for learning in the field of autonomous vehicles and programming can be kept higher~\cite{Medeiros2019}. The idea behind this variation of \textit{competition-based learning} \cite{Burguillo2010} is to have three races in the course that incentivizes the students~\cite{Shim2017} while not using rankings for grading, with the goal to teach more than in comparison to a standard class. Ultimately, in Section \ref{sec_survey}, we ask and answer the following research questions:
\begin{enumerate}
    \item Does the course cover the necessary content to teach autonomous driving?
    \item Is the F1TENTH hardware the right tool to teach autonomous driving hands-on?
    \item Is the aspect of ``racing" a good theme and concept for an educational course?
    \item Is the F1TENTH course helpful for the students' career paths?
\end{enumerate}
In summary, the main contributions of this paper are:

\begin{itemize}
\item A detailed overview of the F1TENTH course, the course philosophy, and the syllabus for teaching autonomous driving with the theme of racing and competition is provided.
\item A modular embedded hardware system called F1TENTH to teach autonomous driving at the university level is explained in detail. This vehicle can run a modular software stack for autonomous driving consisting of modules for perception, planning, and control.
\item Survey results from students that were part of the F1TENTH course are used to answer four defined research questions regarding the quality and necessity of this course.
\end{itemize}

\newpage
\section{Related Work}
\label{sec_related_work}

In this paper, the established term \textit{computational thinking} (CT) is used to describe skills developed by students to solve problems in the field of computer science. Bers et al.~\cite{Bers2014} propose in their research that CT could be increased by leveraging real robots. This is because real hardware (robots) motivates children/students more than only theoretical learning and promotes CT through gamification.

The interest in teaching the fundamentals of autonomous systems and robotics has increased over the last decade. In particular, the number of courses and classes that leverage the usage of real-world robots for teaching and learning different robotic subjects at distinct education levels has grown~\cite{Diago2021}. In a  quasi-experimental study with 24 third-grade students, Diago et al.~\cite{Diago2021} revealed that using educational robotics created statistically significant gains in computational thinking. Especially in contrast to traditional education approaches, the students increased their computational knowledge and achieved a higher level by using real-world robots. Furthermore, by designing hardware-focused and robot-focused courses, students' negative attitudes toward heavy mathematical subjects can be improved~\cite{Kucuk2020}. Since the focus is not only on learning the theory of, e.g., optimization rather than applying software to a robot that moves around, students can develop a higher interest in learning mathematics and further advance their careers in this field~\cite{Kucuk2020}.

Bakala et al.~\cite{Bakala2021} conducted a systematic review of empirical studies that apply classroom robots for preschoolers to enhance computational thinking. Based on reviewing 15 empirical studies, the authors found that mainly commercial robotic kits were used. Unfortunately, in all these kits, only a limited number of input and output interfaces were given, limiting developmental appropriateness to children’s cognitive level. Many studies and evaluations were conducted to get an inside into using robotic hardware kits at the elementary school level~\cite{Jung2018}. In most of these studies, the bee-bot~\cite{Di_Lieto2017} or Lego Mindstorms kits~\cite{Savard2016}~\cite{Irigoyen2013} are used. Only a few publications address the issue of teaching robotics~\cite{Esposito2017} at university (graduate and undergraduate level).

In~\cite{Cielniak2013} the authors describe the educational usage of robotics in an undergraduate Computer Science course. The authors concluded high popularity among the students, a high collaboration within the teams, and high competition in developing individual solutions for the provided robots. Frank et al.~\cite{Frank2018} provide the outline for a first-year design project for autonomous systems. To increase motivation and engagement ~\cite{Medeiros2019}, their project consists of various elements: Team building, team management, budget, document writing, robot design, robot building, and robot programming. Hildebrandt et al.~\cite{Hildebrandt2022} describe in their work the principles and setup of a new course for teaching students the software development for robots. They define five-course principles, after which the course needs to be developed and conclude that using simulation environments for robotic tasks is a huge advantage.

Finally, research was published that explicitly uses autonomous vehicles. \textit{Duckietown}~\cite{Tani2017} teaches students to program a small-scale robot in an urban environment. In~\cite{Feng2018} a scaled RC-Car platform is used to run in a scaled indoor environment, but only a fixed set of hardware and software is provided. The Amazon \emph{DeepRacer}~\cite{Balaji2020} is a small-scale autonomous car used to educate students on simulation and reinforcement learning. Furthermore, international competitions are organized with this vehicle. The most commonly used vehicle is a modified 1:10 scale RC car, and institutions released documentation for hardware and setup on transforming this conventional car into an autonomous racecar. These vehicles are then used either for research or educational purposes and the most prominent one are  the \textit{MIT Racecar}~\cite{Karaman2017}, the \textit{MuSHR racecar}~\cite{srinivasa2019}, the \textit{RoSCAR}~\cite{Hart2014} or the \textit{F1TENTH}~\cite{okelly2019} vehicle.

In summary, the usage of real robots in teaching autonomous systems is applied already. Unfortunately, a course setup that provides applied knowledge for the whole autonomous driving pipeline consisting of hardware selection, software stack development, simulation testing, and real-world application is not available so far. The work presented in this paper builds upon~\cite{okelly2019} and~\cite{Agnihotri2020}. It extends the F1TENTH vehicle in various ways, exploring its capabilities as an educational platform providing now modular hardware and software and ultimately surveying the students about the course's and the vehicles' usefulness.

\section{The F1TENTH Course}
\subsection{Course Philosophy}
The general course philosophy of the proposed course is ``Define the Problem. Implement. Understand'' and  ``Competitions (Races) replace Exams''~\cite{Paul2009}. The goal is to focus on teaching autonomous systems as hands-on as possible with the provided F1TENTH vehicle, allowing students to enhance their computational and systems thinking. With this philosophy, the students learn to think about problems in autonomous driving on their own. The course material is aiming for graduate-student level but can be reduced to undergraduate or even high-school level.

Furthermore, the students participate in three autonomous races with their F1TENTH vehicle during the semester. These races replace the exams in the course. In contrast to competition-based learning, we do not use the ranking of the students in these races for the majority of the grading. The students have to write quality software for the vehicle to be successful in the races. Additionally, the races help the students improve their risk analysis because they must decide how much faster they go with their car to achieve good race results. While the competitive scenario of the races builds up mental toughness for the students, it also creates a way to develop a social community around learning \cite{Papert1980}.

By grouping the students into teams with 2--3 students per team, a diverse set of teams can be created: A mix of majors (only one per team); a mix of programming expertise (Python, C++); a mix of the countries of origin; a mix of genders or ethnic groups. Solving the labs and tasks in these teams improves teamwork and collaboration while enhancing social and emotional learning. Usually, instructors assign the team members after a survey is conducted to gather students' information. Re-teaming can occur once or twice during a 15-week semester.

A final project (or cornerstone project) is set up as an ill-structured software design project with the explicit goal to give the students the experience of struggle and challenge, which can result in failures and setbacks~\cite{Hulls2020}. These failures are intended to teach the student fundamentals of fault diagnosis~\cite{GomezdeGabriel2015} and data visualization. Being guided by the teaching assistants ensures that the project has a reasonable scope. By demonstrating their project at the end of the semester, the students still achieve a positive result and learning outcome~\cite{Hulls2020}. Since no final exam is held, the final grade is composed of the following components and their weighting:

\begin{itemize}
    \item \textbf{40\% Labs:} Results of the code submitted in the different labs.
    \item \textbf{30\% Competition performance:} Results of the races weighted by the race difficulty. 95\% of this grade is based on participation, and only 5\% is based on ranking in the races.
    \item \textbf{20\% Final Project:} Quality of the project demonstration and documentation.
    \item \textbf{5\% Competition document:} An 5-8 page document summarizing the students' approach to the competition (software architecture, algorithms, hardware, tests, etc); examples of performance results, etc.
    \item \textbf{5\% Peer review:} An anonymous evaluation of the students' work performed by their teammates.
\end{itemize}

\subsection{Learning Outcomes}
The learning outcomes for the F1TENTH course were continuously adjusted and refined every semester since the class was first offered in 2018. The course aims to teach autonomous systems hands-on so students can learn about the theoretical software fundamentals of the different autonomy algorithms and apply them to the hardware afterward. The following ten learning outcomes are set up; after the F1TENTH course, the students should be able to

\begin{enumerate}
    \item understand the current challenges in state of the art for autonomous driving,
    \item understand the role of middleware with ROS2 (Robot Operating System 2),
    \item understand common sensors for detection and localization,
    \item explain vehicle dynamic behavior by visualizing vehicle states,
    \item explain the different concepts of path planning,
    \item understand the necessity of stabilizing control actions and the responsibilities of the control algorithm,
    \item design and tune a path tracking controller,
    \item apply software for perception, planning, and control to a 2D and 3D simulation environment,
    \item apply software for perception, planning, and control to the F1TENTH hardware, and
    \item develop their own software for perception, planning and control and apply it to the F1TENTH hardware.
\end{enumerate}

\subsection{Content and Syllabus}

\begin{table}
\caption{F1TENTH Course Syllabus}
\label{tab:course_modules}
\begin{tabular}{cl}
\multicolumn{2}{l}{\cellcolor[HTML]{FFFFFF}{\color[HTML]{000000} \textbf{Module A: Introduction to F1TENTH, the Simulator  \& ROS2}}} \\
{\color[HTML]{000000} 1} & {\color[HTML]{000000} Introduction to Autonomous Driving} \\
{\color[HTML]{000000} 2} & {\color[HTML]{000000} Automatic Emergency Braking} \\
{\color[HTML]{000000} 3} & {\color[HTML]{000000} Rigid Body Transform} \\
\multicolumn{2}{l}{\cellcolor[HTML]{FFFFFF}{\color[HTML]{000000} \textbf{Module B: Reactive Methods}}} \\
{\color[HTML]{000000} 4} & {\color[HTML]{000000} Vehicle States, Vehicle Dynamics and Maps} \\
{\color[HTML]{000000} 5} & {\color[HTML]{000000} Follow the Wall: First Autonomous Drive} \\
{\color[HTML]{000000} 6} & {\color[HTML]{000000} Follow the Gap: Obstacle Avoidance} \\
{\color[HTML]{000000} 7} & {\color[HTML]{000000} Race 1: Preparation} \\
\cellcolor[HTML]{EFEFEF}8 & \cellcolor[HTML]{EFEFEF}Race 1: Single-Vehicle: Obstacle Avoidance\\
\multicolumn{2}{l}{\cellcolor[HTML]{FFFFFF}{\color[HTML]{000000} \textbf{Module C: Mapping \& Localization}}} \\
{\color[HTML]{000000} 9} & {\color[HTML]{000000} Scan matching} \\
{\color[HTML]{000000} 10} & {\color[HTML]{000000} Particle Filter} \\
{\color[HTML]{000000} 11} & {\color[HTML]{000000} Introduction to Graph-based SLAM} \\
\multicolumn{2}{l}{\cellcolor[HTML]{FFFFFF}{\color[HTML]{000000} \textbf{Module D: Planning \& Control}}} \\
{\color[HTML]{000000} 12} & {\color[HTML]{000000} Local Planning: RRT, Spline Based Planner} \\
{\color[HTML]{000000} 13} & {\color[HTML]{000000} Path Tracking: Pure Pursuit} \\
{\color[HTML]{000000} 14} & {\color[HTML]{000000} Path Tracking: Model Predictive Control} \\
{\color[HTML]{000000} 15} & {\color[HTML]{000000} Behavioral Planning: Trustworthy Autonomous Vehicles} \\
\multicolumn{2}{l}{\cellcolor[HTML]{FFFFFF}{\color[HTML]{000000} \textbf{Module E: Vision}}} \\
{\color[HTML]{000000} 16} & {\color[HTML]{000000} Classical Perception: Lane Detection} \\
{\color[HTML]{000000} 17} & {\color[HTML]{000000} Machine Learning Perception: Object Detection} \\
{\color[HTML]{000000} 18} & {\color[HTML]{000000} Final Project Selection} \\
{\color[HTML]{000000} 19} & {\color[HTML]{000000} Race 2: Preparation} \\
\cellcolor[HTML]{EFEFEF}20 & \cellcolor[HTML]{EFEFEF}Race 2: Single-Vehicle: High-Speed \\
\multicolumn{2}{l}{\cellcolor[HTML]{FFFFFF}{\color[HTML]{000000} \textbf{Module F: Special Topics and Invited Talks}}} \\
{\color[HTML]{000000} 21} & {\color[HTML]{000000} Ethics for Autonomous Systems} \\
{\color[HTML]{000000} 22} & {\color[HTML]{000000} Raceline Optimization} \\
{\color[HTML]{000000} 23} & {\color[HTML]{000000} Special Topic 1} \\
{\color[HTML]{000000} 24} & {\color[HTML]{000000} Special Topic 2} \\
{\color[HTML]{000000} 25} & {\color[HTML]{000000} Special Topic 3} \\
\multicolumn{2}{l}{\cellcolor[HTML]{FFFFFF}{\color[HTML]{000000} \textbf{Module G: Race 3 And Project Demonstrations}}} \\
{\color[HTML]{000000} 26} & {\color[HTML]{000000} Race 3: Preparation} \\
\cellcolor[HTML]{EFEFEF}27 & \cellcolor[HTML]{EFEFEF}Race 3: Multi-Vehicle Head-to-Head \\
{\color[HTML]{000000} 28} & {\color[HTML]{000000} Project Demonstrations} \\
\label{tab_syllabus}
\end{tabular}
\end{table}

The F1TENTH course is split into six modules (Module A-F) which consist of 25 lectures (Table \ref{tab_syllabus}). A seventh module (G) is for the final race and the project demonstration. The general idea of the F1TENTH course is to incrementally increase the difficulty of driving with an autonomous vehicle. As depicted in Figure  \ref{fig_course_setup}, the course starts with teaching single-vehicle behavior only and then moves to more complex vehicle behavior like high-speed driving and multi-vehicle scenarios.  Having lab sessions with the real hardware is then providing a concrete learning experience inside the module.


The students learn the foundations of autonomous driving and \textit{Reactive Methods} in Modules A \& B. Here, the car is driving at slow speeds. With the primary sensor (LiDAR)  the students can perceive the environment and avoid obstacles. In the first race, the goal is to drive a single car around a given track while avoiding obstacles. In Modules C \& D, the foundations of localization, planning, and control are explained and a variety of algorithms are presented. The students learn how to localize the car in a given environment with different localization methods. The car's speed is increased by applying different path planning algorithms and path tracking controllers. This part is listed as \textit{high-speed autonomy} and involves heavy tuning since both the localization's accuracy and the controller's quality lead to different vehicle behavior. In the second race, the goal is to drive a single car at high speed around a given track. Finally, in Modules E \& F, object detection is taught, and special topics in autonomous driving are presented. In the final module (G), the students need to apply everything they learned throughout the semester in a multi-vehicle race (2 vehicles against each other) and tune the car to drive fast and reliably. Additionally, the results of the projects are presented.

Teaching autonomous driving is difficult since it involves components from various subfields like computer science, linear algebra, optimization, control theory, and machine learning. As the hypothesis defines that students need to learn autonomous driving in a hardware applied and hands-on way, the course is taught in a \textit{clab} style (classroom + lab). Classes are held twice a week; each session is 80 minutes long (60 minutes lecture + 20 minutes tutorial/lab work). The following core components for the course are established:
\begin{enumerate}
\item \textbf{Theoretical Lectures:} The theoretical fundamentals of the various algorithms in perception, planning, and control are explained in a lecture.

\item \textbf{Labs:} Here the students need to apply the autonomous driving concepts from the lecture to the 2D simulation environment. The labs are explained and discussed in the class, the lab assignments need to be completed outside of class time. The code is evaluated in simulation only.

\item \textbf{Races:} Three races are organized, each testing the students’ application of different class components. The students must bring their code to the car and run either in a single-vehicle or multi-vehicle race.

\item \textbf{Special Topics:} A series of special topics with guest lectures that present their applied autonomous driving work from a research or industry perspective are provided. This gives the students some inspiration about state-of-the-art research and industry work.

\item \textbf{Final Project:} A final project allows the students to propose a problem in autonomous driving and try to solve it using the F1TENTH hardware.
\end{enumerate}


\section{Modular Hardware}
Deploying algorithms on real-world autonomous vehicles is expensive, time-consuming, and dangerous. Especially not many higher education institutions have a real-world autonomous vehicle that can be used for teaching purposes. The primary artifact of this course is an autonomous vehicle called \textit{F1TENTH}. The purpose of the F1TENTH vehicle is to offer a low-cost, low-effort, low-entry bar 1:10 scale vehicle that enables safe and rapid experimentation. In comparison to other educational robots listed in section \ref{sec_related_work}, this small-scale hardware is very close to a real-world vehicle: Ackermann steering; real chassis system with damper and springs; changeable vehicle hardware e.g., tires; different drivetrain setups e.g., AWD and RWD; high-speed and high acceleration. Ultimately, this enables a better understanding of the abstract autonomous driving problems~\cite{Anwar2019} and allows exploring autonomous driving solutions closer to real-world application~\cite{Miller2008}.

\begin{figure}[h]
\begin{center}
\includegraphics[scale=0.93]{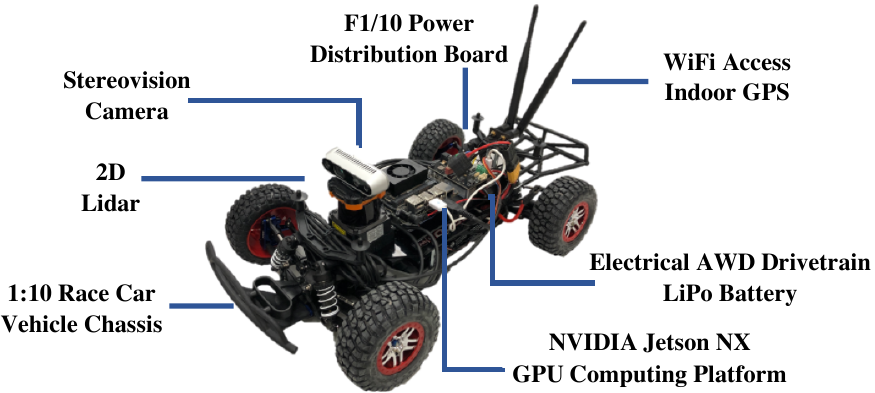}
\caption{F1TENTH hardware setup used in the F1TENTH course.}
\label{fig_f1tenth_hardware}
\end{center}
\end{figure}

As depicted in Figure \ref{fig_f1tenth_hardware}, based on a 1:10 scale remote-controlled vehicle chassis, the car includes various components to transform it into an autonomous vehicle. The car has an electrical all-wheel drivetrain (AWD) powered by a \SI{5000}{\milli\ampere\hour} lithium polymer battery. A specially developed power distribution board powers all electrical components. The F1TENTH vehicle has a 2D LiDAR and a stereovision camera mounted on the front to perceive its environment. The main computation unit is an NVIDIA embedded GPU computer called \textit{Jetson Xavier NX}. This computer has an Ubuntu-based operating system (OS) and allows to use the car like a computer because the keyboard, mouse, and monitor can be plugged in. The course instructors set up the car and handed it out to the students at the beginning of the course.

\begin{figure}[h]
\begin{center}
\includegraphics[scale=0.94]{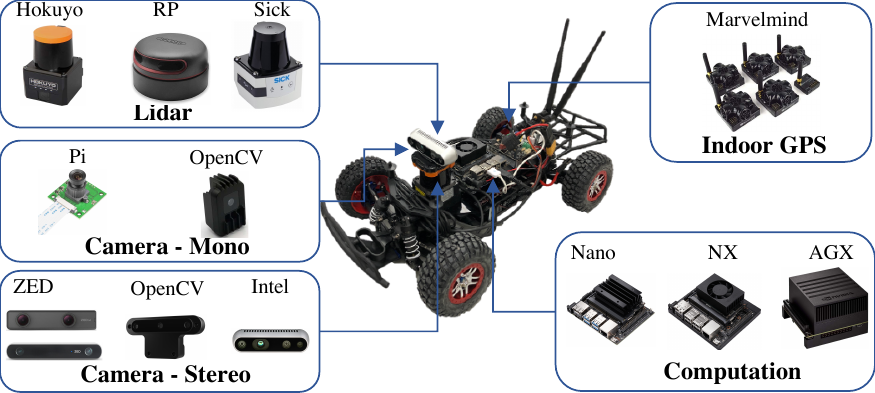}
\caption{F1TENTH hardware: The vehicle offers the possibility of integrating a high variety of different hardware components like camera, LiDAR, or computation systems.}
\label{fig_hardware_comonents}
\end{center}
\end{figure}

All components displayed here are close to setups and parts used in industrial applications. Therefore, working with the car is close to industry standards, and teaching students these systems aims to prepare them explicitly for their first job after the university. Besides this, the F1TENTH car does not have only one fixed hardware setup that is used for teaching. A wide variety of hardware components can be integrated. Figure \ref{fig_hardware_comonents} shows a combination of three LiDARs, two mono cameras, three stereo cameras, and three different computation units that can be used on the F1TENTH vehicle. 

\begin{table*}[]
\caption{Overview of F1TENTH Hardware Modules and their combination for different educational levels}
\label{tab:hardware}

\begin{tabular}{|c|c|cc|ccc|ccc|ccc|}
\hline
\multirow{2}{*}{\textbf{\begin{tabular}[c]{@{}c@{}}Educational\\ Level\end{tabular}}} & \textbf{\begin{tabular}[c]{@{}c@{}}Loca\\ -lization\end{tabular}} & \multicolumn{2}{c|}{\textbf{\begin{tabular}[c]{@{}c@{}}Mono \\ camera\end{tabular}}} & \multicolumn{3}{c|}{\textbf{\begin{tabular}[c]{@{}c@{}}Stereo \\ camera\end{tabular}}} & \multicolumn{3}{c|}{\textbf{\begin{tabular}[c]{@{}c@{}}2D \\ LiDAR\end{tabular}}} & \multicolumn{3}{c|}{\textbf{\begin{tabular}[c]{@{}c@{}}Main \\ Computation Unit\end{tabular}}} \\ \cline{2-13}
 & \begin{tabular}[c]{@{}c@{}}Indoor\\ GPS\end{tabular} & \multicolumn{1}{c|}{\begin{tabular}[c]{@{}c@{}}Raspberry\\ PI\end{tabular}} & \begin{tabular}[c]{@{}c@{}}OpenCV\\ Oak-1\end{tabular} & \multicolumn{1}{c|}{\begin{tabular}[c]{@{}c@{}}Intel\\ Realsense\\ D345i\end{tabular}} & \multicolumn{1}{c|}{\begin{tabular}[c]{@{}c@{}}Zed Mini\\ Zed \\ ZED2\end{tabular}} & \begin{tabular}[c]{@{}c@{}}OpenCV\\ Oak-D\end{tabular} & \multicolumn{1}{c|}{\begin{tabular}[c]{@{}c@{}}Hokuyo\\ 10LX\\ 30LX\end{tabular}} & \multicolumn{1}{c|}{Sick} & \begin{tabular}[c]{@{}c@{}}RP\\ A3M1\end{tabular} & \multicolumn{1}{c|}{\begin{tabular}[c]{@{}c@{}}Nvidia\\ Jetson\\ Nano\end{tabular}} & \multicolumn{1}{c|}{\begin{tabular}[c]{@{}c@{}}Nvidia\\ Jetson\\ NX\end{tabular}} & \begin{tabular}[c]{@{}c@{}}Nvidia\\ Jetson\\ AGX\end{tabular} \\ \hline
\textbf{\begin{tabular}[c]{@{}c@{}}High \\ School\end{tabular}} &  & \multicolumn{1}{c|}{X} &  & \multicolumn{1}{c|}{} & \multicolumn{1}{c|}{} &  & \multicolumn{1}{c|}{} & \multicolumn{1}{c|}{} &  & \multicolumn{1}{c|}{X} & \multicolumn{1}{c|}{} &  \\ \hline
\textbf{\begin{tabular}[c]{@{}c@{}}University:\\ Undergraduate\end{tabular}} &  & \multicolumn{1}{c|}{} & X & \multicolumn{1}{c|}{} & \multicolumn{1}{c|}{} &  & \multicolumn{1}{c|}{} & \multicolumn{1}{c|}{} & X & \multicolumn{1}{c|}{} & \multicolumn{1}{c|}{X} &  \\ \hline
\textbf{\begin{tabular}[c]{@{}c@{}}University:\\ Graduate\end{tabular}} &  & \multicolumn{1}{c|}{} & X & \multicolumn{1}{c|}{X} & \multicolumn{1}{c|}{} & X & \multicolumn{1}{c|}{} & \multicolumn{1}{c|}{X} &  & \multicolumn{1}{c|}{} & \multicolumn{1}{c|}{X} &  \\ \hline
\textbf{\begin{tabular}[c]{@{}c@{}}Research and  \\Industry \\ Training \end{tabular}} & X & \multicolumn{1}{c|}{} &  & \multicolumn{1}{c|}{} & \multicolumn{1}{c|}{X} &  & \multicolumn{1}{c|}{X} & \multicolumn{1}{c|}{} &  & \multicolumn{1}{c|}{} & \multicolumn{1}{c|}{} & X \\ \hline

\end{tabular}
\end{table*}

This modular hardware setup of the F1TENTH vehicle provides many advantages. First, the possibility of switching to a different sensor component is given. For example, the 2D LiDAR sensors offer different sampling rates, field-of-views, and ranges. Changing them on the vehicle leads to an impact on the autonomy software e.g., obtaining distance data of obstacles. Another example is the usage of different computation hardware. By running e.g., the same path planner and controller on other computation hardware, the students experience slower/faster algorithm calculation times, leading to a slower/faster control frequency and ultimately to worse/better car control. The NVIDIA Jetson computer differs in the overall performance (TOPS, TFLOPS), the number of GPU and CPU cores, RAM memory, and SSD storage~\cite{nvidiajetson}.
Second, based on the set of heuristics for the development of educational robots defined by~\cite{Giang2019} all 14 defined heuristics are fulfilled. These include a high level of \textit{adaptability}, the possibility for \textit{collaboration and communication}, the \textit{relevance} of the autonomous driving task and the list of \textit{challenges} provided for the students throughout the course.
Third, the modular hardware leads to the possibility of using the F1TENTH for a wider variety of teaching purposes on different educational levels. In discussions and interviews with former students and other F1TENTH teachers, four hardware setups for different educational levels displayed in Table \ref{tab:hardware} were defined.

\section{Modular Software Stack}
\subsection{F1TENTH Stack}

While the hardware provides the fundamentals of an autonomous vehicle and creates the constraints of what the system can do, the software is the part where the real magic of an autonomous vehicle happens. Everything the human does typically in a conventional vehicle must be incorporated into the autonomous driving software. The complexity increases with each level of automation - more profound knowledge and even more software is required.

\begin{figure}[h]
\begin{center}
\includegraphics[scale=0.95]{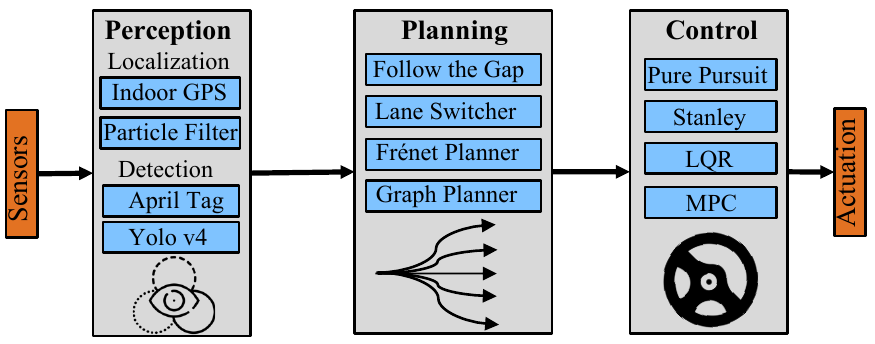}
\caption{Software Modules for the F1TENTH vehicle}
\label{fig_software_stack}
\end{center}
\end{figure}

In an autonomous vehicle, many software components need to be combined; this is usually called a \textit{software stack}. For autonomous vehicles, this software stack consists of the three big modules \textit{perception, planning, and control} which will enable safe and robust autonomous operation in real-world situations~\cite{Levinson2011}. For the F1TENTH vehicle and the course, a completely new software stack was developed, consisting of the following software modules displayed in Figure \ref{fig_software_stack}.

\begin{figure}[h]
\begin{center}
\includegraphics[scale=0.95]{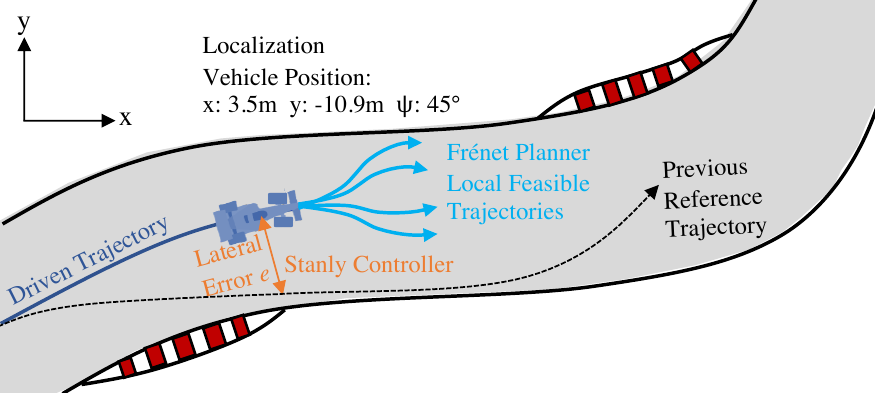}
\caption{Example of a combination of software modules from the F1TENTH software stack. Localization: GPS; Planning: Frenet Planner; Control: Stanley}
\label{fig_software_combination}
\end{center}
\end{figure}

Although many examples of software stacks are given in research papers, the center of attention in this project was to create a \textit{modular} software stack. The goal was to include as many different software modules as possible to offer the students a wide variety of algorithms. First, this allows teaching simpler algorithms at the course's beginning and moving on to more difficult algorithms later. Second, this enables comparing the quality of all algorithms. For example, all algorithms in the control module can track a predefined path, but some algorithms achieve a better tracking quality than others. Third, all algorithms have a different need for computation power and need to apply resources on either the GPU or CPU. Fourth, since not all perception, planning, and control algorithms fit well together, this modularity enables the demonstration of coupled effects between the individual modules.
Exemplary, Figure \ref{fig_software_combination} shows the combination of three algorithms from the F1TENTH stack. While the vehicle receives its current pose (localization), it is trying to track a previously calculated reference trajectory (control). The vehicle is constantly generating new feasible trajectories (planning) to find the optimal path without hitting an obstacle.

The perception modules consist of algorithms from the field of localization and detection. With these algorithms, the autonomous vehicle can find its position and heading (pose) and detect obstacles (e.g. other vehicles) in front of it. The perception modules in the F1TENTH stack consist of the following algorithms.
\begin{itemize}
    \item \textbf{Localization - Indoor GPS:} The car can get its absolute position from and indoor GPS hardware. This hardware is localizing via triangulation and sends an absolute position in a pre-defined area with an accuracy of about 2~cm. In the simulation, the GPS position is provided with absolute ground truth.

    \item \textbf{Localization - Particle Filter:}  A particle filer is a localization algorithm that uses a set of random positional samples that update every detection to approximate the car's position. When equipped with a LiDAR, the vehicle can run a particle filter to localize based on LiDAR point cloud detection and a map of the environment. The map needs to be created beforehand with the LiDAR.

    \item \textbf{Detection - AprilTag:} An AprilTag is a set of 2D barcodes designed to be detected quickly and accurately. With camera calibration, AprilTag detection will provide encoded information and the relative translation and rotation between the camera and the opponent car.

    \item \textbf{Detection - YOLO v4 Object Detection:} For more advanced teaching, deep-learning-based object detection is used. A simple neural network structure called YOLO~~\cite{Redmon2016} is used, which takes in camera images and outputs bounding box detection. Positional information can be calculated based on camera calibration. Students can use a pre-trained neural network or explore their own designs of neural networks to perform object detection
\end{itemize}

The planning module consists of algorithms that plan a trajectory in front of the vehicle. A trajectory consists of a path (x- and y-Position) and a velocity profile. The trajectories need to be collision-free and enable a feasible vehicle behavior. The planning modules in the F1TENTH stack consist of the following algorithms.
\begin{itemize}
    \item \textbf{Gap Follower} finds gaps in the LiDAR scan by finding the broadest range of scan angles with the highest depth value~~\cite{sezer2012novel}. Then this algorithm steers the vehicle to follow the most significant gap to avoid obstacles.
	\item \textbf{Lane Switcher} creates equispaced lanes that span the entire track and utilize an optimal curvature race line~~\cite{Heilmeier2019}. The algorithm switches to a specific lane or back to the race line when trying to overtake or block an opponent.
	\item \textbf{Frenet Planner} is based on a semi-reactive method~~\cite{Werling2010}. This planner can select goal coordinates in the Frenet-Frame of the racetrack and generate multiple trajectories to follow the optimal race line and avoid obstacles by choosing the appropriate trajectory.
	\item \textbf{Graph Planner} generates a graph covering the race track~~\cite{Stahl2019}. The nodes in the graph are vehicle poses in the world frame, and the edges of the graph are generated trajectories similar to those in the Frenet Planner. The algorithm selects appropriate actions for the vehicle from the action set for overtaking and following when traversing the graph.
\end{itemize}

Finally, the control module includes all algorithms that track the desired path and velocity of the planned path. The control modules consist of the following algorithms.
\begin{itemize}
    \item \textbf{Pure Pursuit} is a geometric path tracker~\cite{coulter1992implementation}. Using a fixed distance look-ahead point on the planned path (reference), a steering angle can be calculated, making the vehicle steer correctly on the path.
    \item \textbf{Stanley} is a geometric path tracker~\cite{thrun2006stanley}. Here, the goal is to minimize the heading and cross-track errors (deviation from the reference trajectory). A correction steering angle can be calculated based on both errors.
    \item \textbf{LQR (Linear Quadratic Regulator)} is an optimization-based technique. The LQR is reducing the lateral error from the reference path and is optimizing regarding a given cost function. The output is an optimal vehicle speed and steering.
    \item \textbf{MPC (Model Predictive Control)} is an optimization-based technique~\cite{Katriniok2013}. The MPC looks at a given receding horizon into the future, predicts the vehicle behavior (vehicle states) for these time steps, and then solves an optimization problem based on constraints. The output is optimal vehicle acceleration and steering.
\end{itemize}

The modular software leads to the possibility of using the F1TENTH for a wider variety of teaching purposes on different educational levels. In discussions and interviews with former students and other F1TENTH teachers, four software setups for different educational levels displayed in Table \ref{tab:hardware} were defined.


\begin{table*}[]
\centering
\caption{Overview of F1TENTH Software Modules and their combination for different educational levels}
\label{tab:software}
\begin{tabular}{|c|cc|cccc|cccc|}

\hline
\multirow{2}{*}{\textbf{\begin{tabular}[c]{@{}c@{}}Educational \\ Level\end{tabular}}} & \multicolumn{2}{c|}{\textbf{Perception}} & \multicolumn{4}{c|}{\textbf{Planning}} & \multicolumn{4}{c|}{\textbf{Control}} \\ \cline{2-11}
 & \multicolumn{1}{c|}{Localization} & Detection & \multicolumn{1}{c|}{\begin{tabular}[c]{@{}c@{}}Follow\\ the\\ Gap\end{tabular}} & \multicolumn{1}{c|}{\begin{tabular}[c]{@{}c@{}}Lane\\ Switcher\end{tabular}} & \multicolumn{1}{c|}{\begin{tabular}[c]{@{}c@{}}Frenet \\ Planner\end{tabular}} & \begin{tabular}[c]{@{}c@{}}Graph\\ Planner\end{tabular} & \multicolumn{1}{c|}{\begin{tabular}[c]{@{}c@{}}Pure\\ Pursuit\end{tabular}} & \multicolumn{1}{c|}{Stanley} & \multicolumn{1}{c|}{LQR} & MPC \\ \hline
\textbf{High School} & \multicolumn{1}{c|}{-} &  & \multicolumn{1}{c|}{X} & \multicolumn{1}{c|}{} & \multicolumn{1}{c|}{} &  & \multicolumn{1}{c|}{X} & \multicolumn{1}{c|}{} & \multicolumn{1}{c|}{} &  \\ \hline
\textbf{\begin{tabular}[c]{@{}c@{}}University:\\ Undergraduate\end{tabular}} & \multicolumn{1}{c|}{-} &  & \multicolumn{1}{c|}{X} & \multicolumn{1}{c|}{} & \multicolumn{1}{c|}{} &  & \multicolumn{1}{c|}{X} & \multicolumn{1}{c|}{X} & \multicolumn{1}{c|}{} &  \\ \hline
\textbf{\begin{tabular}[c]{@{}c@{}}University:\\ Graduate\end{tabular}} & \multicolumn{1}{c|}{\begin{tabular}[c]{@{}c@{}}GPS\\ Particle Filter\end{tabular}} & Yolo v4 & \multicolumn{1}{c|}{X} & \multicolumn{1}{c|}{X} & \multicolumn{1}{c|}{X} &  & \multicolumn{1}{c|}{X} & \multicolumn{1}{c|}{X} & \multicolumn{1}{c|}{X} &  \\ \hline
\textbf{\begin{tabular}[c]{@{}c@{}}Research and  \\Industry \\ Training\end{tabular}} & \multicolumn{1}{c|}{\begin{tabular}[c]{@{}c@{}}GPS\\ Particle Filter\end{tabular}} & Yolo v4 & \multicolumn{1}{c|}{} & \multicolumn{1}{c|}{} & \multicolumn{1}{c|}{X} & X & \multicolumn{1}{c|}{} & \multicolumn{1}{c|}{} & \multicolumn{1}{c|}{X} & X \\ \hline
\end{tabular}
\end{table*}

\subsection{Autoware Auto}
The largest open-source project and community around software and hardware development for self-driving vehicles is the so-called \textit{Autoware Foundation (AWF)}. The autonomous driving software developed by AWF is currently used in a wide variety of real-world use-cases by external users across SAE-level 4 automated person transportation platform, high-speed autonomous driving on the racetrack, valet parking, logistics and material handling for movement of goods. The F1TENTH vehicle also runs AWFs open-source software stack called \textit{Autoware}~\cite{Kato2015}. This software stack is built on ROS2 and consists of all the functionality required for autonomous driving, i.e., software for perception, planning, and control. All components are developed in a modular architecture with crisply defined interfaces and APIs.

By bringing Autoware to the F1TENTH platform, students can develop on a small-scale platform and leverage significant portions of a real-world autonomous vehicle software stack as they move to full-scale vehicles. Since the F1TENTH stack is tailor-made for the usage on the F1TENTH car and even specifies the racing purpose, the Autoware software leverages the real-world application of an autonomous vehicle much more. Since the hurdle for understanding and running Autoware on the F1TENTH vehicle is much higher, the usage of Autoware is only taught only at the end of the semester in a special session.

\section{Simulation Environments}
Simulating the behavior of autonomous vehicles is a crucial element in development and education. While simulations are used in R\&D to ensure the safety and maturity of the algorithm, in education, it is used to teach the proposed software components in a safe and reliable environment. One source of failures is dismissed by excluding the hardware, focusing on teaching the algorithm fundamentals and educating the students using software-in-the-loop (SiL) environments. While a variety of simulation environments and platforms for autonomous vehicles exists~\cite{Yang2021}, this course offers two simulation environments.

\subsection{2D-Simulator}
\begin{figure}[h]
\begin{center}
\includegraphics[scale=0.95]{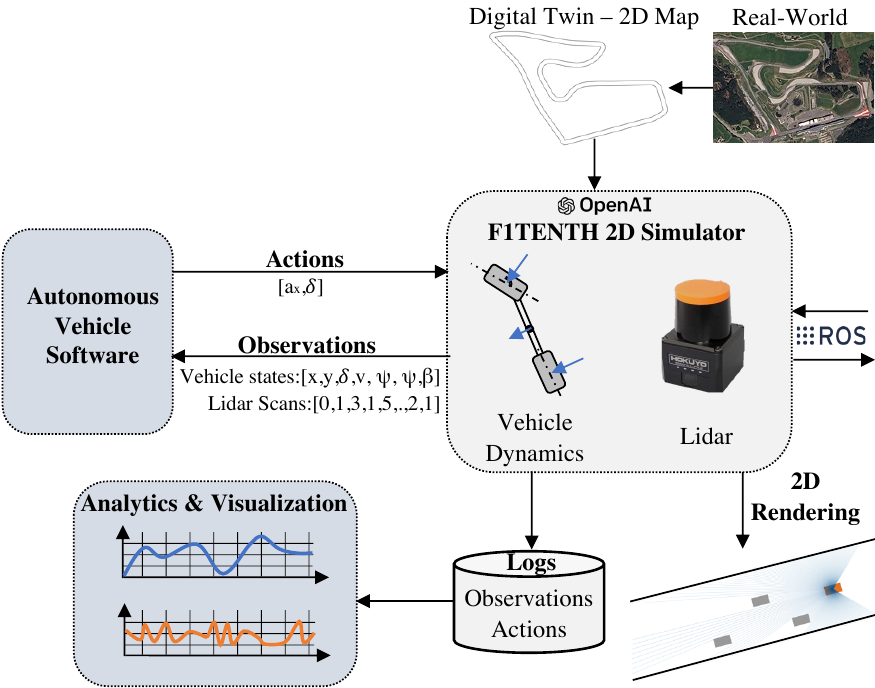}
\caption{Process and workflow of the F1TENTH 2D Simulator}
\label{fig_F1TENTHgym}
\end{center}
\end{figure}

For fast evaluation and testing of the code developed by the students, a 2D simulation environment is provided. This is based on an open-source 2D autonomous vehicle simulator created to test path planners and controllers for the F1TENTH autonomous racing competition~~\cite{okelly2019}. The advantage of this simulator is that it is lightweight and runs on all OS (Mac, Linux, Windows). The students can run their developed code directly without any significant changes. The simulation environment is set up in Python code and enables the exchange with ROS2 via an additional bridge. Figure \ref{fig_F1TENTHgym} shows the process and workflow of the F1TENTH 2D simulator and the related components.

The 2D environment is deterministic with realistic vehicle dynamics based on a single-track vehicle dynamics model~~\cite{Althoff2017}. This means that the vehicle maneuvers are closer to the physical limits and significant effects like understeering and oversteering are simulated with a linear tire forces approximation. In addition, collision with the racetrack boundaries and other vehicles is detected automatically, giving the students feedback that their code failed. Additionally, a 2D LiDAR sensor simulation is integrated. This enables simple perception-based algorithms such as object detection (clustering) and localization methods.

The physics engine used in this simulator is faster than real-time simulation and state serialization (loading and saving), making this simulation environment interesting for running experimental evaluations simultaneously. We provide an additional tool that allows the students to visualize the collected data in the simulator. This is necessary to gain insights into vehicle behavior to debug the developed autonomous driving software. Furthermore, this simulator has an Open AI Gym~\cite{brockman2016openai} interface, enabling further education in the field of reinforcement learning.

\begin{figure}[h]
\begin{center}
\includegraphics[scale=0.15]{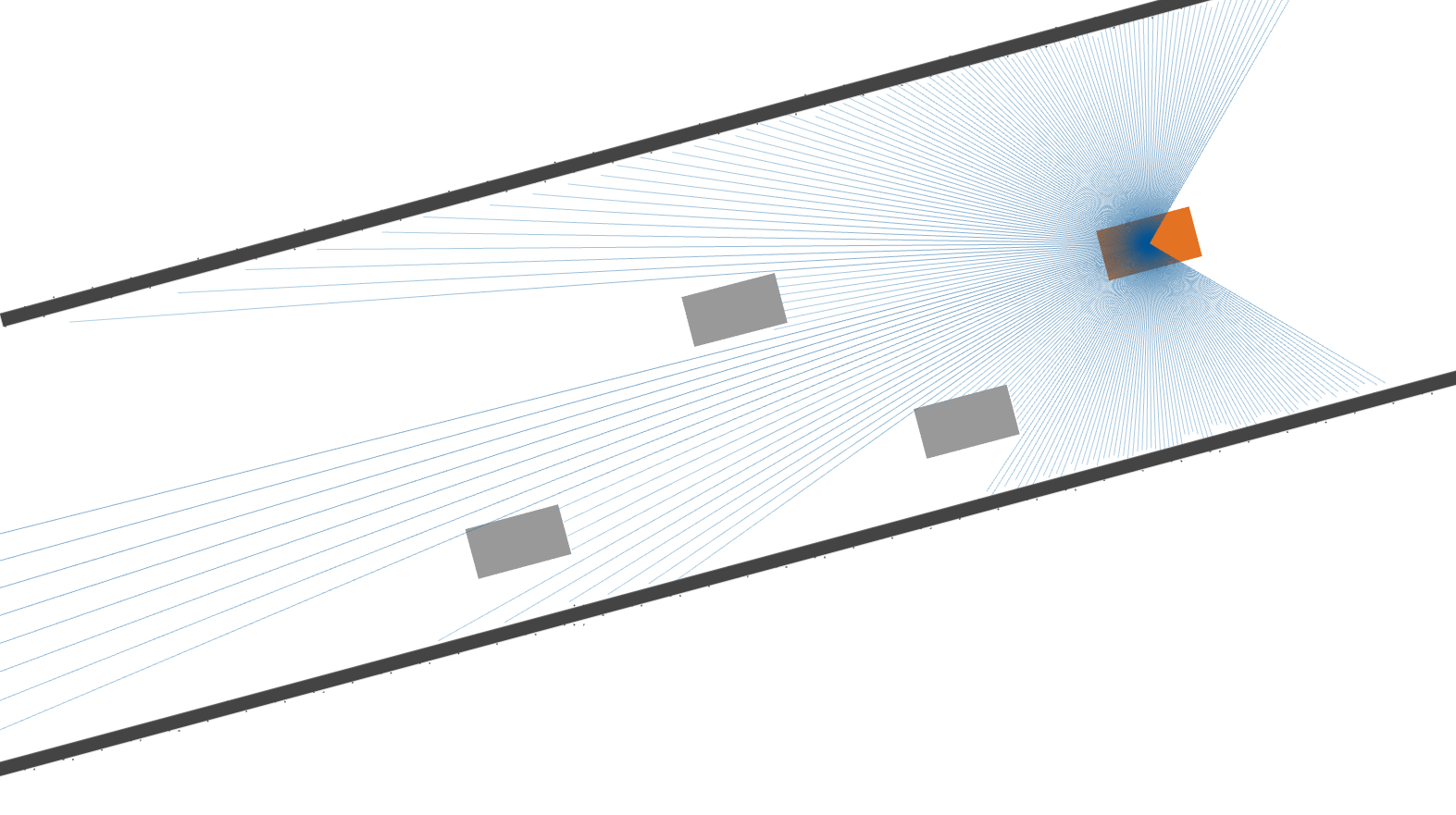}
\caption{Exemplary 2D rendering of the F1TENTH Simulator. The ego vehicle is depicted as the orange box, driving in a multi-vehicle (grey boxes) environment. The ego vehicle visualizes its LiDAR stream (blue lines), which detects the obstacles and the walls along the track.}
\label{fig_F1TENTHgym2}
\end{center}
\end{figure}

Finally, the simulation environment is modular and allows different racetracks to be integrated. In the current course setup over 20 real-world racetracks are provided as a digital twin 2D map for the simulation. With various maps, students can test their algorithms on different racetrack layouts: High-speed turns, narrow hairpins, or slow sweeper curves. Figure \ref{fig_F1TENTHgym2} shows an exemplary 2D rendering of the simulator in a multi-vehicle environment.

\subsection{3D-Simulator}
Since the 2D simulator is a lightweight environment that focuses more on planning and control, an additional 3D simulator is introduced: The \textit{SVL Simulator}~\cite{Guodong2020} is an open-source end-to-end simulation platform for autonomous vehicles. The vehicles and the environment are modeled in 3D (Unity engine) and provide photo-realistic and physically correct object simulation (e.g., crashes). Furthermore, the simulator offers a wide variety of sensor plugins with additional noise to be more realistic. This simulator is used to teach object and lane detection with the camera.

\begin{figure}[h]
\begin{center}
\includegraphics[scale=0.23]{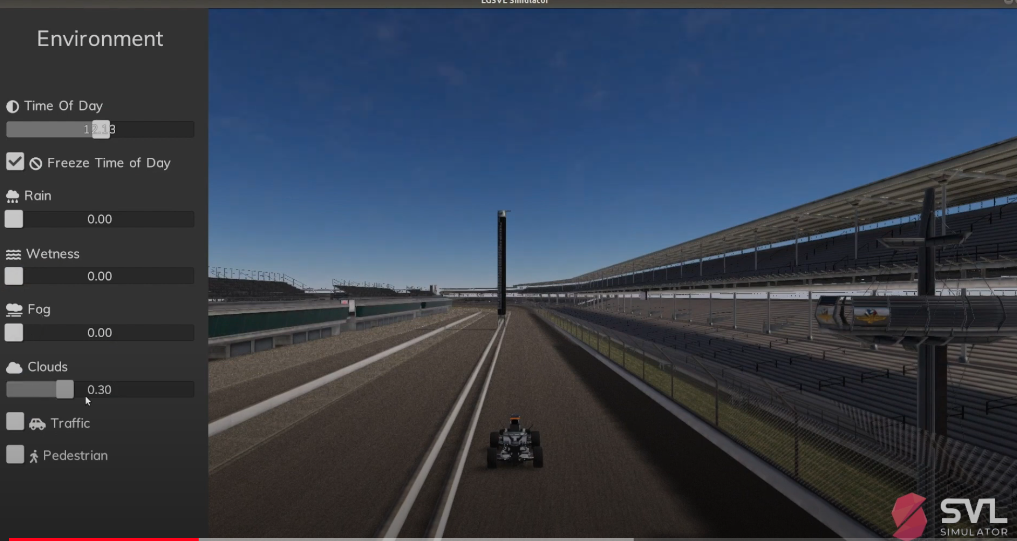}
\caption{F1TENTH vehicle in the SVL simulator~\cite{Guodong2020} on a 1:10 scale version on a 3D racetrack}
\label{fig_SVL}
\end{center}
\end{figure}

The SVL Simulator was mainly built for engineers to verify and test their software stack for autonomous vehicles. Therefore, this simulator requires heavy computation resources (GPU, CPU, RAM), which may be inaccessible to all students and their laptops. Since SVL offers a cloud simulation, the goal is to bring the SVL simulator to a cloud environment in the future, so the students can access those on the cloud instead of using their computers. Figure \ref{fig_SVL} displays an exemplary simulation setup in the SVL simulator with the F1TENTH vehicle on the racetrack.

\section{Course Survey}
\label{sec_survey}
For this paper, a formal assessment was conducted in terms of a student survey at the semester end of spring 2022. The survey was handed out to the students after finishing all mandatory work but before the final course grade was given. The course survey was done at four universities (4-year R-1 institutions?) that taught the course in spring 2022, reaching 48 students in total. The survey was anonymous so no demographic data (gender, ethnicity) was collected this time. To exclude the bias created by various instructors, teaching styles, and university setups, the focus of the survey questions primarily assesses the usage of the vehicle and the theme of "racing" in the course. To condense and structure the survey results, four research questions are defined that are answered with the help of the survey outcomes.

\subsection{Q1: Does the course cover the necessary content to teach autonomous driving?}
\begin{figure}[h]
\begin{center}
\includegraphics[scale=0.95]{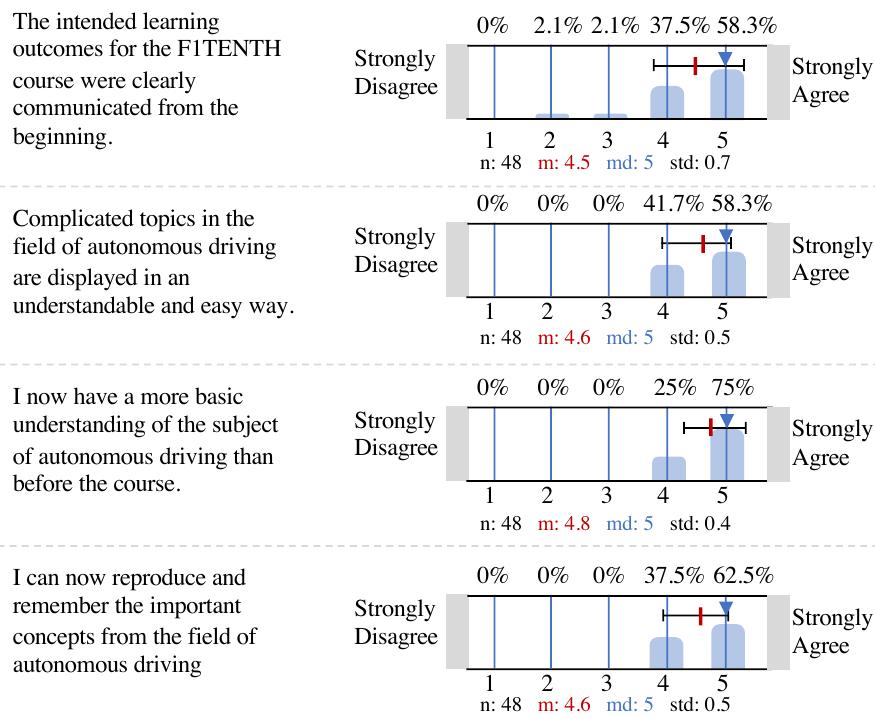}
\caption{Survey Results: These questions are related to the content of the F1TENTH lecture. The students give feedback on the quality of the lecture content and display if the lecture material is thorough enough for teaching autonomous driving content (n: number of answers, m: mean value, md: median value, std: standard deviation)}
\label{survey_results1}
\end{center}
\end{figure}

First, the students are surveyed regarding the course content and if, from the student's perspective, the topic of autonomous driving is covered holistically in the course. Survey questions and results related to this research question are displayed in Figure \ref{survey_results1}.

The results show that more than half of the students strongly agree that the learning outcomes were communicated clearly and that the complicated topics were displayed understandably. Since these questions scored the lowest in this subsection, we conclude that the course content needs to be simplified in some lectures. Although it is evident that this course is teaching the complete pipeline of autonomous driving, we see here potential to make the material more apparent to the students. Exactly 75\% of the surveyed students strongly believe they now have a more fundamental understanding of autonomous driving technology. The answer supports the feedback that 62.5\% of the students can reproduce the most important concepts (perception, planning, control) from the field of autonomous driving. This indicates that the course content contains all essential aspects of teaching autonomous driving.
\subsection{Q2: Is the F1TENTH hardware the right tool to teach autonomous driving hands-on?}
\begin{figure}[h]
\begin{center}
\includegraphics[scale=0.95]{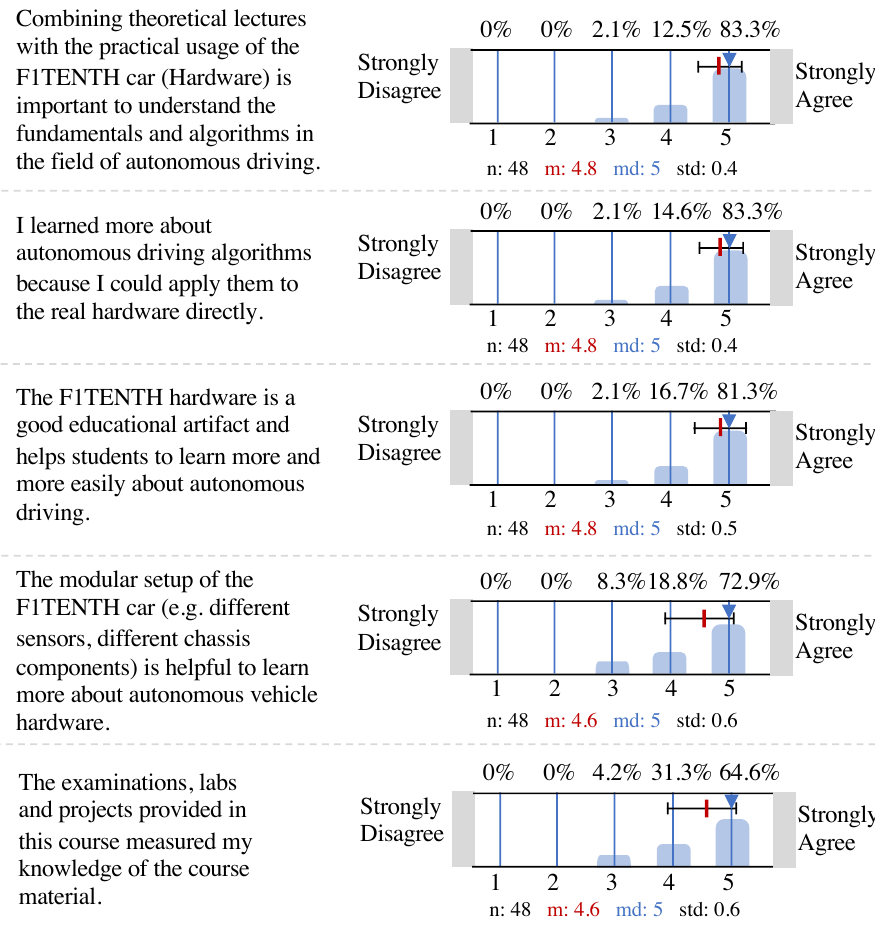}
\caption{Survey Results: These questions are related to the usage of the F1TENTH Hardware in the course. The students give feedback on the usage of the hardware and if they think this vehicle helped them to learn more about the topic. (n: number of answers, m: mean value, md: median value, std: standard deviation)}
\label{survey_results2}
\end{center}
\end{figure}

Second, the students are surveyed regarding the usage of the F1TENTH hardware. The goal is to get feedback on whether the small-scale vehicle is good support for teaching autonomous driving-related topics and if it helps the students learn content in this area. Survey questions and results related to this research question  are displayed in Figure \ref{survey_results2}
With a mean value of 4.8, the students strongly agree that combining theory and real hardware leads to a better learning outcome in the field of autonomous driving. Additionally, 82.3\% of the students think the F1TENTH vehicle is a good educational artifact. This reveals that the proposed F1TENTH hardware has a high educational value for learning about autonomous driving. Although generally, the students strongly agree that the modular setup of the F1TENTH car is helpful, it does not seem as valuable as the general vehicle hardware itself. Additionally, as general feedback, the students answered the questions "What did you most like about the course?" with the following written answers: Applying the code to real hardware; Working with the car; Cars and Hardware; Working with Hardware; The hands-on work and the competitive spirit of the course.

\subsection{Q3: Is the aspect of ``racing" a good theme and concept for an educational course? }
\begin{figure}[h]
\begin{center}
\includegraphics[scale=0.95]{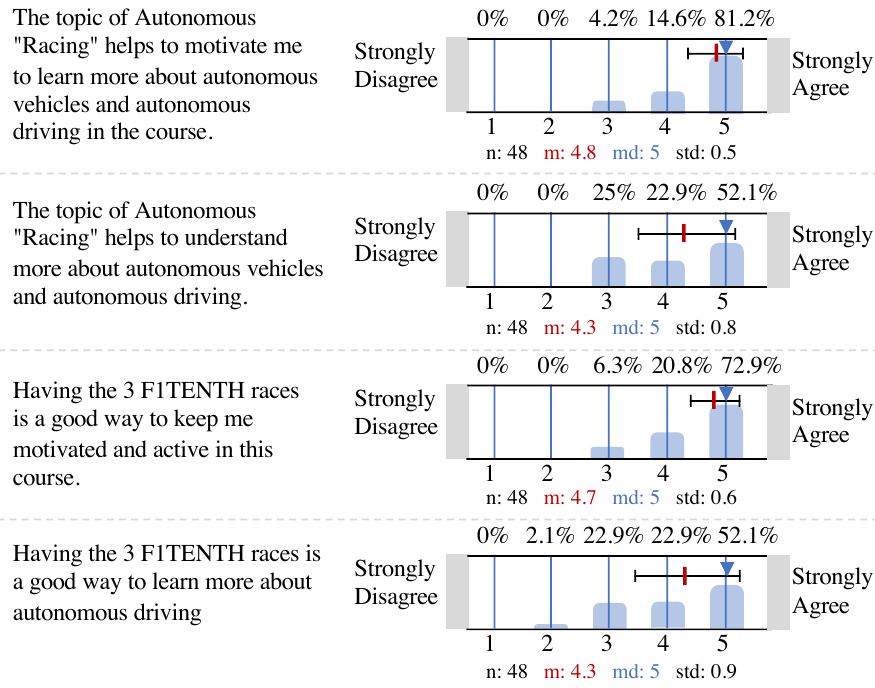}
\caption{Survey Results: These questions are related to the racing theme of the F1TENTH course and the three competitions throughout the semester. (n: number of answers, m: mean value, md: median value, std: standard deviation)}
\label{survey_results3}
\end{center}
\end{figure}

Third, the students are surveyed regarding the course's racing theme and their thoughts on competing in three different races with the F1TENTH vehicle. Survey questions and results related to this research question are displayed in Figure \ref{survey_results3}.

The goal was to identify if the students felt that both the topic of racing and the three races helped them stay motivated and active in the course. With an average answer of 4.8 and 4.7 to these questions, the students indicated that they strongly agree that the racing setup helps them to stay motivated. We conclude that the students acknowledged the general course philosophy "Competitions replace Exams" in a way that they liked to come to the lecture and stayed motivated throughout the whole semester. In addition, we wanted to know if the students feel that both the topic of racing and the three races help them understand and learn more about autonomous driving. These questions were received with an average of 4.3 --- the lowest score in the survey. Only 50\% of the students strongly agree that the topic of racing, albeit fun and motivating, has added value in helping them learn the subject matter. The general observation was that the racing tracks and rules often lead to more complex vehicle behaviors that the lectures may not clearly explain. Also, high speeds and high accelerations, which are often needed for winning the races, usually favor simpler algorithms in the reactive paradigm rather than more advanced planning algorithms. Winning a racing also calls for extensive trial-and-error testing, which is a niche application/aspect often ignored by most college courses and can be regarded as repetitive without additional educational value.

\subsection{Q4: Is the F1TENTH course helpful for the students' career paths? }
\begin{figure}[h]
\begin{center}
\includegraphics[scale=0.95]{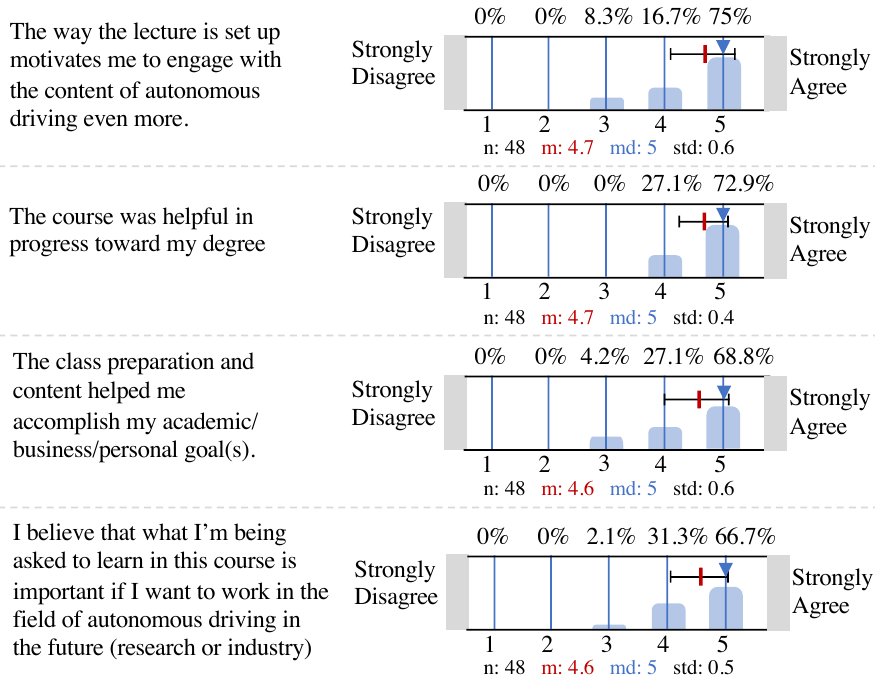}
\caption{Survey Results: These questions are related to the career perspectives of the students. (n: number of answers, m: mean value, md: median value, std: standard deviation)}
\label{survey_results4}
\end{center}
\end{figure}

In the final questionnaires, the students are asked whether the course was relevant to their future career plans and if this course would help them get more involved in autonomous driving. Survey questions and results related to this research question  are  displayed in Figure \ref{survey_results4}.

Over 70\% of the students indicated that this course was helpful in their progress toward their degree. Additionally, 75\% of the students strongly agree that the F1TENTH lectures motivated them to engage more with the topic of autonomous driving. Additionally, 66.7\% of the students strongly agree that the course teaches them important content if they want to work in the field of autonomous driving in the future. It can be concluded that the course setup provides the proper range to teach future students the essential topics in the field of autonomous driving. Furthermore, the course content provides all the necessary know-how to provide the students for their first job after the university - either in research or academia.

\section{Research and Industry}

The F1TENTH vehicle is around \$3500 in hardware costs making it affordable and a low barrier to entry for real-world autonomous vehicle development and testing. The F1TENTH vehicle and software can be used effectively in a research context - either in academia to answer fundamental research questions or as a training tool in the industry. Current approaches for autonomous driving, especially in industry, involve comprehensive testing in simulation coupled with extensive track testing to provide sufficient coverage of canonical and edge cases. Simulation allows for quick and efficient coverage of a variety of scenarios at the expense of some realism. In contrast, track testing allows for realism at the cost of sparse test coverage due to resource limitations (track time, drone vehicles, pedestrian simulators, etc). This additional intermediate testing stage combines more realism than simulation but less infrastructure than full-scale track testing. This will accelerate autonomous vehicle development and increase safety for engineers.

The small-scale hardware is portable and easy to maintain, set up, and tear down, allowing for simplified operation and testing almost anywhere without extensive site preparations. Since the F1TENTH vehicle has a good representation of the dynamics of a full-scale vehicle, the F1TENTH vehicle is utilized to apply new planning and control algorithms~\cite{Bulsara2020,Ivanov2020,jain2020, Tatulea2020, Rosolia2020, Pagot2020}. For example, in~\cite{Brunnbauer2019}, and~\cite{Gotlib2019}, the vehicle is used to demonstrate new localization techniques by focusing on detecting cones or walls.

As a low-risk platform, the F1TENTH vehicle is ideal for performing research requiring real-world dynamics that are dangerous to test with full-scale vehicles, such as high-speed off-road driving~\cite{xiao2021ikd,karnan2022viikd}. In addition, the platform's low cost also makes it particularly conducive to multi-agent controls research, where it is feasible to demonstrate state-of-the-art distributed controller synthesis on multiple F1TENTH cars~\cite{wei2021onevision}.
Although these are small-scale vehicles, the cars achieve high speeds ($\approx$60 km/h) and accelerations ($\approx$15 m/s$^{2}$) for their size. These vehicles' results demonstrate real-world race cars' characteristics, such as changing tire dynamics based on temperature and wear changes during driving, making them effective analog testing tools aiming toward full-scale autonomous racing.

Finally, there is an effort to utilize small-scale vehicles for testing next-generation advanced driver assistance systems (ADAS) in the industry. The current state of ADAS in commercially available vehicles consists of a collection of individual ADAS functionalities operating in parallel. Complex interactions between these systems will emerge as this collection grows in scope and capabilities, especially in edge cases. While automated driving design and verification have strong foundations stemming from robotics or aerospace validation, this is insufficient when considering human-centered driving approaches and scaled vehicles offer a complementary testing and development tool. 

\section{Discussion and Conclusions}
The goal of the F1TENTH course is to teach autonomous systems more hands-on and to combine the theory of perception, planning, and control with the application of the learned content on an actual autonomous vehicle. The F1TENTH car fulfills these requirements by offering a 1:10 small-scale platform to the students. By keeping it the hardware and software close to industry practices the course creates a sense of authenticity among learners and support them better to engage with questions regarding software and hardware development for autonomous systems. The survey results confirm the hypothesis that the F1TENTH vehicle is an excellent educational artifact and helps students to learn more and more efficiently about autonomous driving.

Both the displayed hardware and the software stack provide a highly modular setup to teach content in the field of autonomous driving. This modularity and variety of hardware and software were not provided by any other course yet. This setup allows the teacher to teach various autonomous driving content in perception, planning, and control. As a significant advantage, the modular hardware will enable teachers to teach autonomous systems at different educational levels. On the downside, currently, the F1TENTH hardware has high costs (around 5000 USD), which leads to the fact that not all universities can afford ten or more cars for their students. In the future, the aim is to reduce the costs of vehicle hardware to create a more economic solution that can be taken up by a broader variety of schools, not just the resourceful ones.

The course survey gives valuable feedback on the course structure. The students indicated that the racing theme motivates them to stay active in the course, but it has no significant learning advantage. We conclude that new theme formats with the vehicle can be explored, for example, a \textit{cargo delivery} theme or a \textit{valet parking} theme. Since this course was taught in 2018 for the first time, there are five generations of former F1TENTH students. For future educational research, we want to survey the former F1TENTH students and learn more about the impact of the F1TENTH course on their careers.

Since all course material, hardware setup instructions and the software stack are made open-source, the adaptation rate of the F1TENTH course by various universities is much higher. The co-authors of this paper are all instructors of a version of the F1TENTH course at their home universities and adapted the usage of the F1TENTH vehicle as part of their teaching. Additionally, regularly international competitions are offered so students from different universities can meet each other at conferences and compete with their F1TENTH vehicles. We also consider these conferences and exhibition fairs instead of competitions  because being competitive usually biases towards drawing more males or other dominant groups into the activity, i.e., students who are certain of their abilities to compete and potentially win.

Furthermore, the F1TENTH hardware is currently used in various research projects to evaluate the next generation of autonomy algorithms. This provides the chance to use the F1TENTH vehicle beyond higher education systems and use it even in companies that want to teach or onboard new employees in the field of autonomous driving.

\section{Summary}
In this paper, an autonomous vehicle hardware platform is presented that was developed for teaching autonomous systems hands-on. This article describes the course syllabus and teaching modules with the goal of hands-on teaching of autonomous driving fundamentals. The course increases the difficulty by introducing more advanced software and using more hardware components. From basic reactive methods to advanced planning algorithms, the teaching labs enhance students' computational thinking and systems thinking through autonomous driving with the F1TENTH vehicle.

The hardware setup of the F1TENTH vehicle and the software stack for the vehicle and course are explained in detail.  Both modular setups allow to teach autonomous driving on different educational levels: Simple hardware and algorithms in the beginning and more complex hardware and algorithms in the end. With this design, it is possible to teach the theoretical fundamentals and apply them to real-world hardware. The goal is that students learn the critical aspect of using, tuning, and combining software modules with real autonomous hardware. The F1TENTH vehicle focuses on teaching university students the fundamentals of autonomous driving, but the modular hardware allows the usage on many educational levels to teach autonomous systems.

Ultimately, the student's feedback in the survey indicated that the course provides the right content to teach autonomous driving. More than 80\% of the students strongly agree that the hardware platform and lab modules greatly motivate their learning, and more than 70\% of the students strongly agree that the hardware-enhanced their understanding of the subjects. Based on this finding it can be concluded that teaching autonomous driving in theory and combining it with the hands-on application of the car is essential to learning more about autonomous systems. In contrast, survey results on racing competitions show that, although more than 80\% of the students strongly agree that the competitions motivate them and only 50\% strongly agree that the competitions enhance their learning outcomes. Therefore the topic of racing is an excellent way to motivate the students throughout the semester, but it has no additional educational value.

Furthermore, all work presented in this paper is made available open-source and interested students, teachers and researchers can access the \href{https://courses.f1tenth.org/}{F1TENTH course material on openEdx}, \href{https://f1tenth.org/build.html}{F1TENTH hardware build}, \href{https://github.com/f1tenth/f1tenth_gym}{F1TENTH 2D Simulator} and the \href{https://github.com/f1tenth}{F1TENTH software stack} online.

\section*{Contributions and Acknowledgement}
Johannes Betz initiated the idea of this paper. Johannes Betz and Hongrui Zheng are the main developers and maintainers of both the modular hardware and software of the vehicle and were TAs in the F1TENTH class. Zirui Zang and Florian Sauerbeck contributed to hardware and software development. Zirui Zang was also a TA in the F1TENTH class. Rosa Zheng, Joydeep Biswas, Krzystof Walas, Madhur Behl, and Venkat Krovi contributed to the syllabus setup, taught the course at their universities, and contributed to the survey. Velin Dimitrov contributed to the research section. Rahul Mangharam contributed to the overall structure of the paper and the course syllabus and revised the paper critically. \\


%

\bibliographystyle{IEEEtran}
\bibliography{bibliography.bib}

\vspace{-20pt}

\begin{IEEEbiography}[{\includegraphics[width=1in,height=1.25in,clip,keepaspectratio]{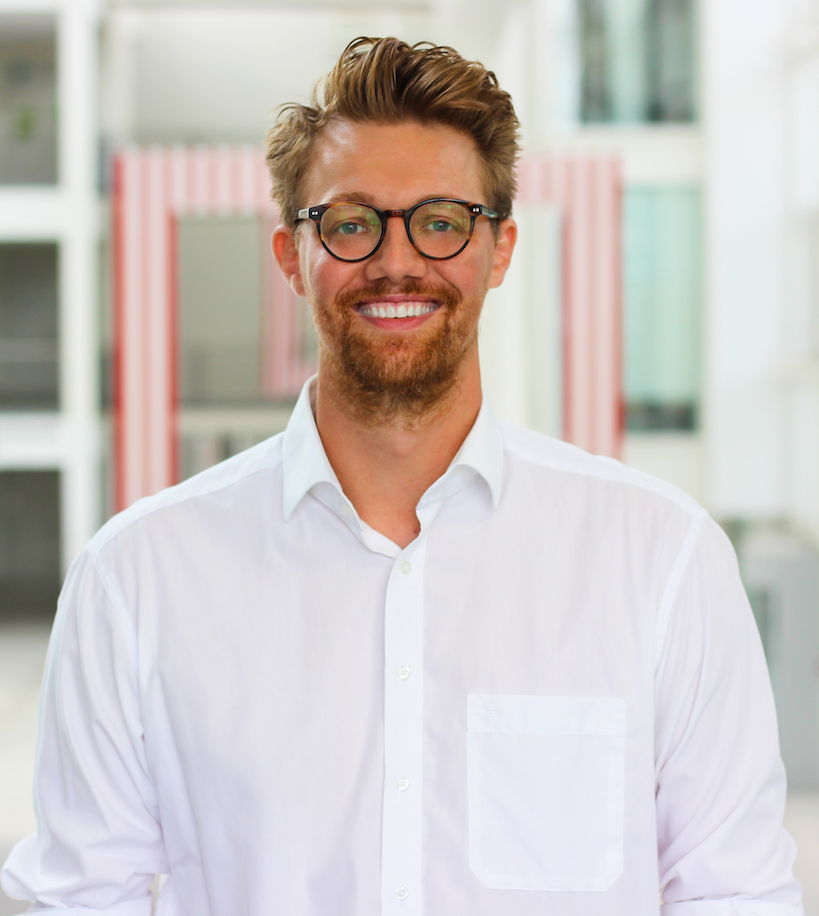}}]
{Johannes Betz} earned both a B. Eng. (2011) and a M. Sc. (2012) in the field of Automotive Engineering. After he did is PhD at the Technical University of Munich (TUM) he was Postdoctoral Researcher at the Institute of Automotive Technology at TUM where he founded the TUM Autonomous Motorsport Team. He is now a postdoctoral researcher at the University of Pennsylvania where he is working at the xLab for Safe Autonomous Systems. His research is focusing on a holistic software development for autonomous systems with extreme motions at the dynamic limits in extreme and unknown environments. 
\end{IEEEbiography}

\vspace{-20pt}

\begin{IEEEbiography}[{\includegraphics[width=1in,height=1.25in,clip,keepaspectratio]{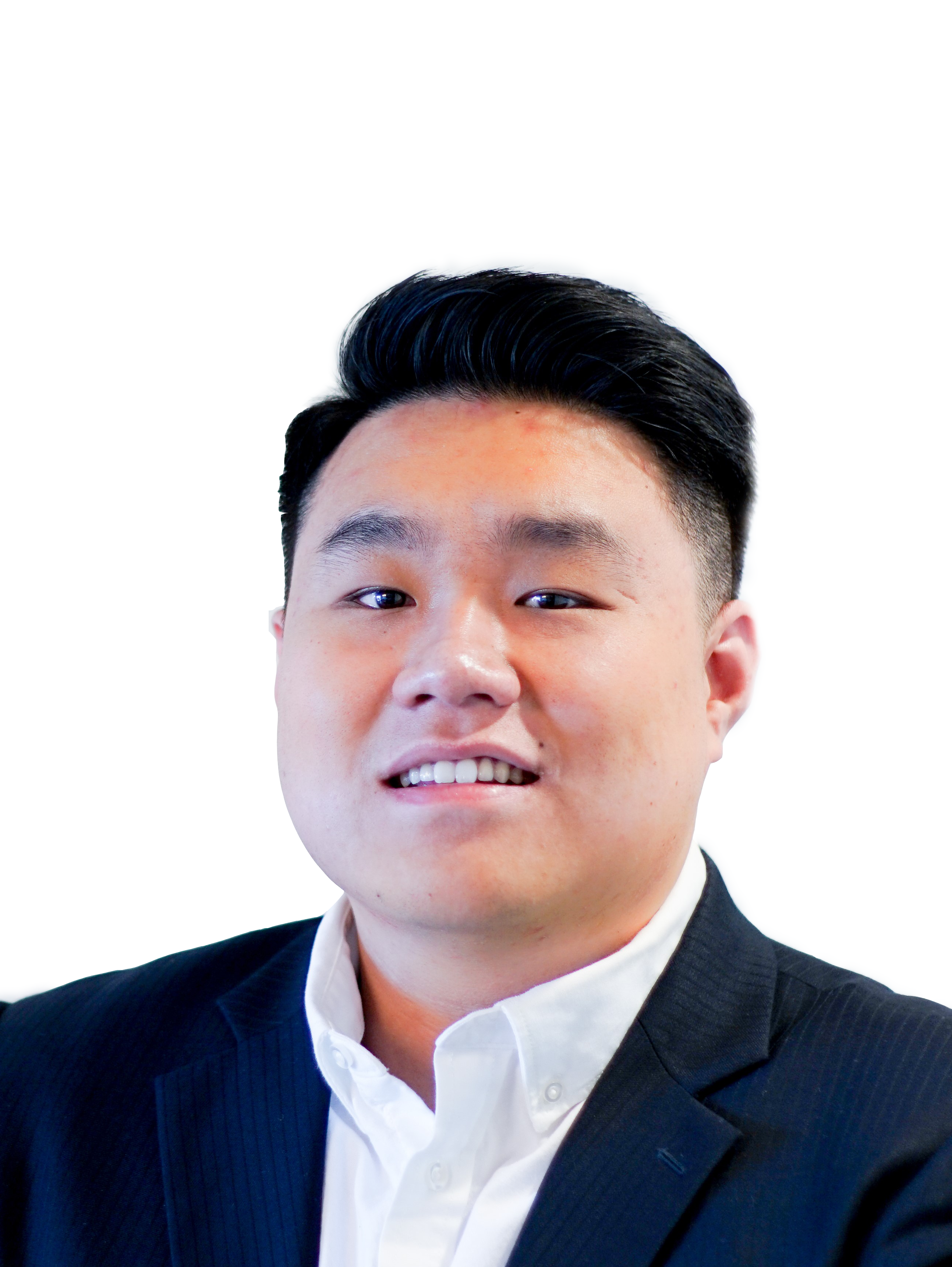}}]
{Hongrui Zheng} received his B.S. degrees in Mechanical Engineering and Computer Science from Georgia Institute of Technology, and his M.S. degree in Robotics from University of Pennsylvania. He is currently a Ph.D. candidate at the University of Pennsylvania where he is working at the xLab for Safe Autonomous Systems. His research focuses on building the tools and theoretical foundations necessary to scale design, testing, and optimization of safe-autonomous systems.
\end{IEEEbiography}
\vspace{-20pt}
\begin{IEEEbiography}[{\includegraphics[width=1in,height=1.25in,clip,keepaspectratio]{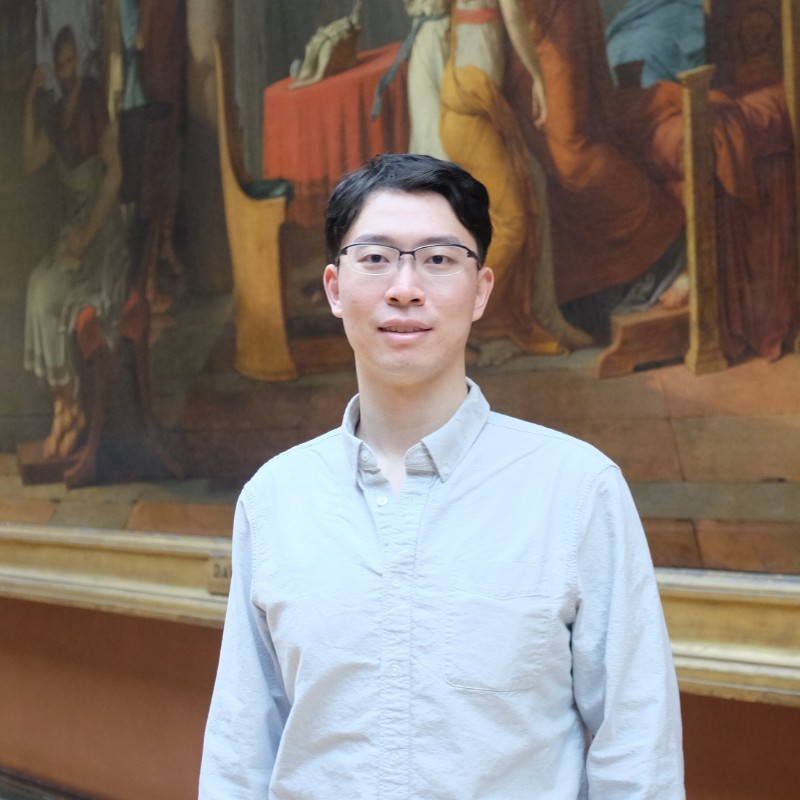}}]
{Zirui Zang} received his B.S. degree in Optical Engineering from University of Rochester and his M.S. degree in Robotics from University of Pennsylvania. He is currently a Ph.D. candidate at the University of Pennsylvania where he is working at the xLab for Safe Autonomous Systems. His research focuses on multi-agent interactions between autonomous vehicles.
\end{IEEEbiography}
\vspace{-20pt}
\begin{IEEEbiography}[{\includegraphics[width=1in,height=1.25in,clip,keepaspectratio]{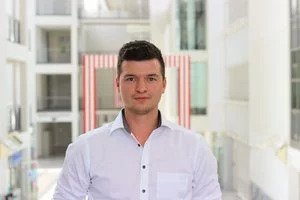}}]
{Florian Sauerbeck} is currently a PhD student at the Technical University of Munich. He received both his B.Sc. degree in 2017 and his M.Sc. degree in 2019 from the Technical University of Munich (TUM), Munich, Germany, where he is currently working towards his Ph.D. in mechanical engineering at the Institute of Automotive Technology. His main research interests include localization, object detection and sensor fusion related applications for autonomous driving with the focus on unstructured environments.
\end{IEEEbiography}
\vspace{-20pt}
\begin{IEEEbiography}[{\includegraphics[width=1in,height=1.25in,clip,keepaspectratio]{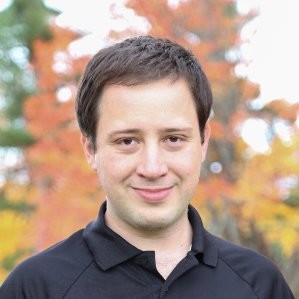}}]
{Velin Dimitrov} is a Senior Research Engineer at Toyota Research Institute. Prior to that he was a Postdoctoral Research Associate in the Electrical and Computer Engineering department at Northeastern University. Velin received his Bachelors of Science in Electrical and Computer Engineering from Franklin W. Olin College of Engineering, his Masters of Science in Robotics Engineering from Worcester Polytechnic Institute, and PhD in Electrical and Computer Engineering from Northeastern University. Velin's areas of interest include human-in-the-loop shared control and assistive robotics.
\end{IEEEbiography}
\vspace{-20pt}
\begin{IEEEbiography}
[{\includegraphics[width=1in,height=1.25in,clip,keepaspectratio]{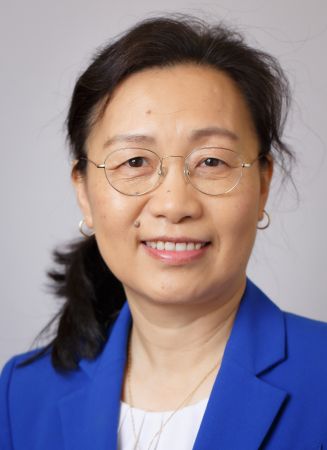}}]
{Yahong Rosa Zheng} received the B.S. degree from the University of Electronic Science and Technology of China, Chengdu, China, in 1987, and the M.S. degree from Tsinghua University, Beijing, China, in 1989, both in electrical engineering. She received the Ph.D. degree from the Department of Systems and Computer Engineering, Carleton University, Ottawa, Canada, in 2002. Currently she is a full professor a Lehigh University, her research interests include array signal processing, wireless communications, and wireless sensor network. 
\end{IEEEbiography}
\vspace{-15pt}
\begin{IEEEbiography}[{\includegraphics[width=1in,height=1.25in,clip,keepaspectratio]{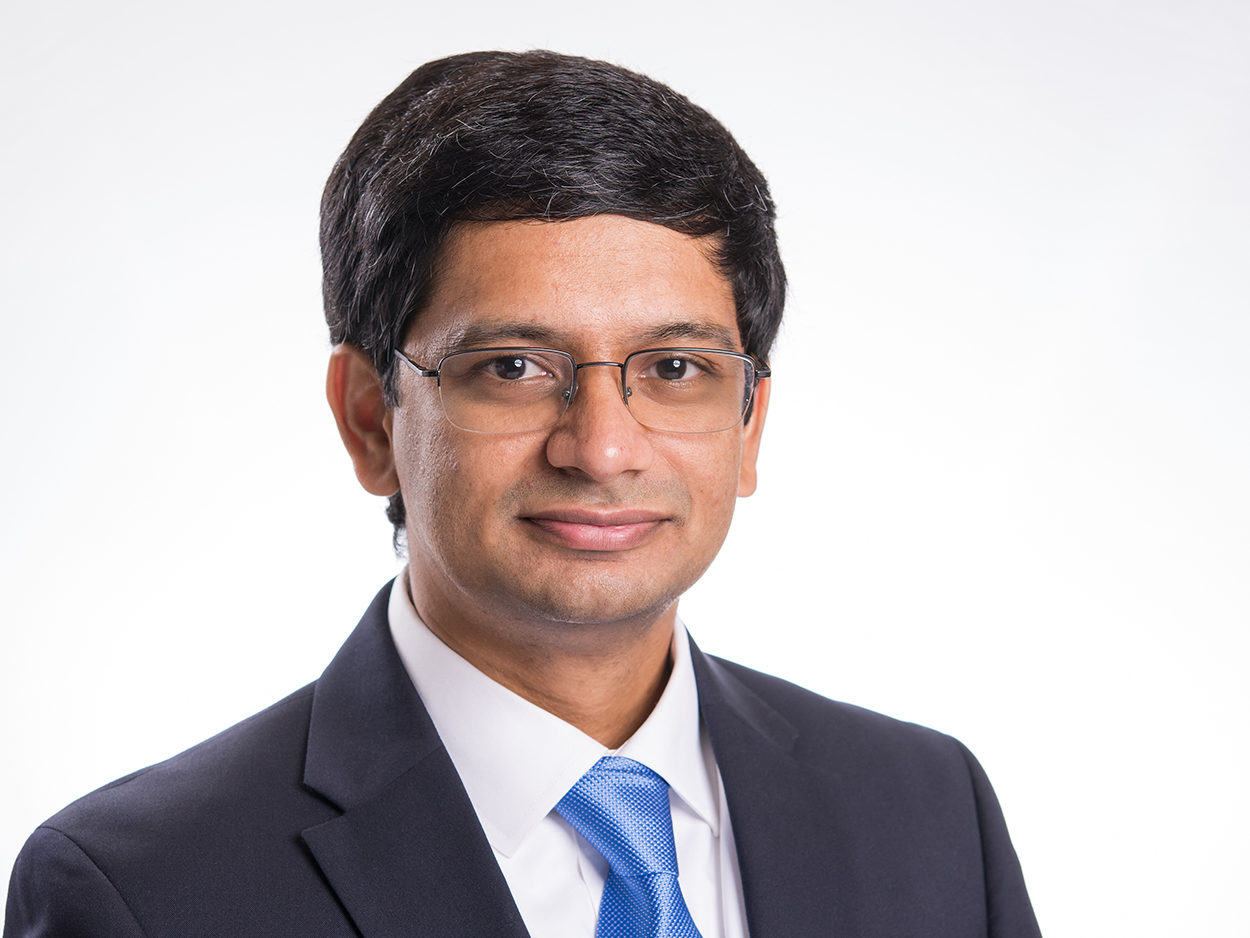}}]
{Joydeep Biswas'} earned his PhD in Robotics from Carnegie Mellon University in 2014, and his B.Tech in Engineering Physics from the Indian Institute of Technology Bombay in 2008. He was an Assistant Professor in the College of Information and Computer Sciences at University of Massachusetts Amherst. Currently Joydeep is an Assistant Professor at UT Austin.

\end{IEEEbiography}
\vspace{-15pt}
\begin{IEEEbiography}[{\includegraphics[width=1in,height=1.25in,clip,keepaspectratio]{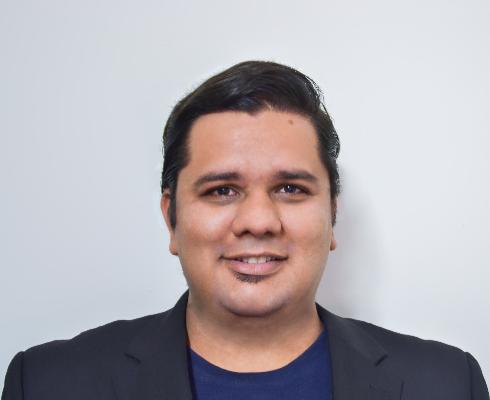}}]
{Madhur Behl} is an assistant professor in the departments of Computer Science, and Systems and Information Engineering, and a member of the Cyber-Physical Systems Link Lab at the University of Virginia.  He conducts research at the confluence of Machine Learning, Predictive Control, and Artificial Intelligence with applications in Cyber-Physical Systems, Autonomous Systems, Robotics, and Smart Cities. He received his Ph.D. (2015) and M.S. (2012), in Electrical and Systems Engineering, both from the University of Pennsylvania.
\end{IEEEbiography}
\vspace{-15pt}
\begin{IEEEbiography}[{\includegraphics[width=1in,height=1.25in,clip,keepaspectratio]{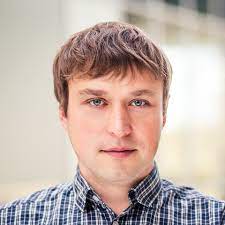}}]
{ Krzystof Walas} graduated from Poznan University of Technology (PUT) in Poland receiving MSc in
Automatic Control and Robotics. He received (with honours) Ph.D. in Robotics in 2012 for his thesis concerning legged robots locomotion in structured environments. Currently, he is an Assistant Professor in the Institute of Robotics and Machine Intelligence at PUT, Poland. His research interests are related to robotic perception for physical interaction applied both to walking and grasping tasks.
\end{IEEEbiography}
\vspace{-15pt}
\begin{IEEEbiography}[{\includegraphics[width=1in,height=1.25in,clip,keepaspectratio]{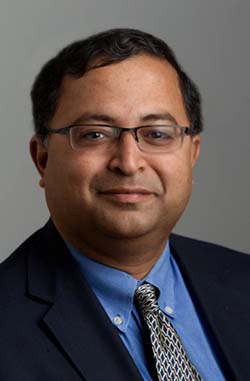}}]
{Venkat Krovi} is the Michelin Endowed Chair of Vehicle Automation in the Departments of Automotive Engineering and Mechanical Engineering at Clemson University. He received his Ph.D. degree in Mechanical Engineering and Applied Mechanics from the University of Pennsylvania in 1998.
\end{IEEEbiography}
\vspace{-15pt}
\begin{IEEEbiography}[{\includegraphics[width=1in,height=1.25in,clip,keepaspectratio]{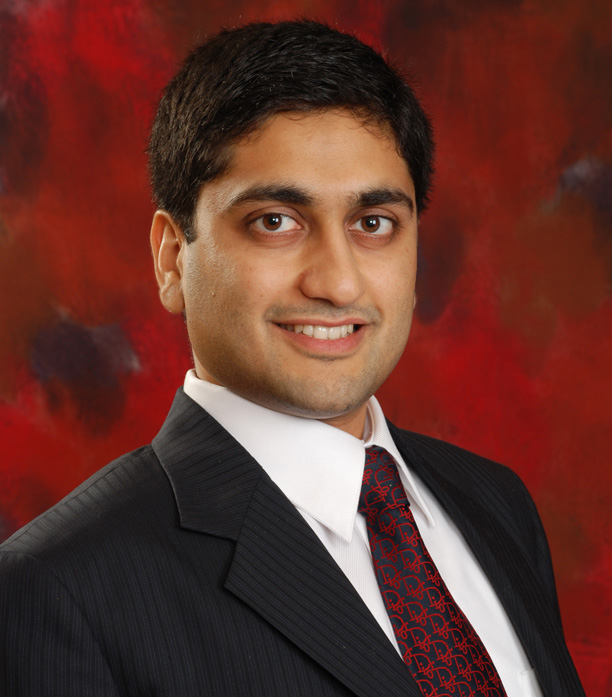}}]
{Rahul Mangharam} received his Ph.D. in Electrical and Computer Engineering from Carnegie Mellon University where he also received his MS and BS.  He is an Associate Professor in the Department of Electrical and Systems Engineering at the University of Pennsylvania. He is a founding member of the PRECISE Center and directs the xLab for Safe Autonomous Systems Lab at Penn. His research is at the intersection of formal methods,  machine learning and control for medical devices,  energy efficient buildings, and autonomous systems.
\end{IEEEbiography}

\vfill

\end{document}

%% file: ORCIDheader.tex
\usepackage{scalerel}
\usepackage{tikz}
\usetikzlibrary{svg.path}
\definecolor{orcidlogocol}{HTML}{A6CE39}
\tikzset{
  orcidlogo/.pic={
    \fill[orcidlogocol] svg{M256,128c0,70.7-57.3,128-128,128C57.3,256,0,198.7,0,128C0,57.3,57.3,0,128,0C198.7,0,256,57.3,256,128z};
    \fill[white] svg{M86.3,186.2H70.9V79.1h15.4v48.4V186.2z}
                 svg{M108.9,79.1h41.6c39.6,0,57,28.3,57,53.6c0,27.5-21.5,53.6-56.8,53.6h-41.8V79.1z M124.3,172.4h24.5c34.9,0,42.9-26.5,42.9-39.7c0-21.5-13.7-39.7-43.7-39.7h-23.7V172.4z}
                 svg{M88.7,56.8c0,5.5-4.5,10.1-10.1,10.1c-5.6,0-10.1-4.6-10.1-10.1c0-5.6,4.5-10.1,10.1-10.1C84.2,46.7,88.7,51.3,88.7,56.8z};
  }
}
\newcommand\orcidicon[1]{\href{https://orcid.org/#1}{\mbox{\scalerel*{
\begin{tikzpicture}[yscale=-1,transform shape]
\pic{orcidlogo};
\end{tikzpicture}
}{|}}}}

\usepackage{hyperref} 

%% file: main.bbl
\begin{thebibliography}{10}
\providecommand{\url}[1]{#1}
\csname url@samestyle\endcsname
\providecommand{\newblock}{\relax}
\providecommand{\bibinfo}[2]{#2}
\providecommand{\BIBentrySTDinterwordspacing}{\spaceskip=0pt\relax}
\providecommand{\BIBentryALTinterwordstretchfactor}{4}
\providecommand{\BIBentryALTinterwordspacing}{\spaceskip=\fontdimen2\font plus
\BIBentryALTinterwordstretchfactor\fontdimen3\font minus
  \fontdimen4\font\relax}
\providecommand{\BIBforeignlanguage}[2]{{%
\expandafter\ifx\csname l@#1\endcsname\relax
\typeout{** WARNING: IEEEtran.bst: No hyphenation pattern has been}%
\typeout{** loaded for the language `#1'. Using the pattern for}%
\typeout{** the default language instead.}%
\else
\language=\csname l@#1\endcsname
\fi
#2}}
\providecommand{\BIBdecl}{\relax}
\BIBdecl

\bibitem{Pribyl2019}
O.~Pribyl and M.~Lom, ``Impact of autonomous vehicles in cities: User
  perception,'' in \emph{2019 Smart City Symposium Prague ({SCSP})}.\hskip 1em
  plus 0.5em minus 0.4em\relax IEEE, May 2019.

\bibitem{Winkle2016}
\BIBentryALTinterwordspacing
T.~Winkle, ``Safety benefits of automated vehicles: Extended findings from
  accident research for development, validation and testing,'' in
  \emph{Autonomous Driving}.\hskip 1em plus 0.5em minus 0.4em\relax Springer
  Berlin Heidelberg, 2016, pp. 335--364. [Online]. Available:
  \url{https://doi.org/10.1007/978-3-662-48847-8_17}
\BIBentrySTDinterwordspacing

\bibitem{hayes_2020}
\BIBentryALTinterwordspacing
A.~Hayes, ``Self-driving cars could change the auto industry (gm, f),'' Aug
  2020. [Online]. Available:
  \url{https://www.investopedia.com/articles/personal-finance/031315/selfdriving-cars-could-change-auto-industry.asp}
\BIBentrySTDinterwordspacing

\bibitem{Wadud2016}
Z.~Wadud, D.~MacKenzie, and P.~Leiby, ``Help or hindrance? the travel, energy
  and carbon impacts of highly automated vehicles,'' \emph{Transportation
  Research Part A: Policy and Practice}, vol.~86, pp. 1--18, Apr. 2016.

\bibitem{pendlton2017}
\BIBentryALTinterwordspacing
S.~D. Pendleton, H.~Andersen, X.~Du, X.~Shen, M.~Meghjani, Y.~H. Eng, D.~Rus,
  and M.~H. Ang, ``Perception, planning, control, and coordination for
  autonomous vehicles,'' \emph{Machines}, vol.~5, no.~1, 2017. [Online].
  Available: \url{https://www.mdpi.com/2075-1702/5/1/6}
\BIBentrySTDinterwordspacing

\bibitem{Renz2021}
A.~Renz and R.~Hilbig, ``\BIBforeignlanguage{en}{Correction to: Prerequisites
  for artificial intelligence in further education: identification of drivers,
  barriers, and business models of educational technology companies},''
  \emph{\BIBforeignlanguage{en}{Int. J. Educ. Technol. High. Educ.}}, vol.~18,
  no.~1, Dec. 2021.

\bibitem{Tang2018}
\BIBentryALTinterwordspacing
J.~Tang, L.~Shaoshan, S.~Pei, S.~Zuckerman, L.~Chen, W.~Shi, and J.-L. Gaudiot,
  ``Teaching autonomous driving using a modular and integrated approach,'' in
  \emph{2018 {IEEE} 42nd Annual Computer Software and Applications Conference
  ({COMPSAC})}.\hskip 1em plus 0.5em minus 0.4em\relax {IEEE}, Jul. 2018.
  [Online]. Available: \url{https://doi.org/10.1109/compsac.2018.00057}
\BIBentrySTDinterwordspacing

\bibitem{Shibata2021}
\BIBentryALTinterwordspacing
M.~Shibata, K.~Demura, S.~Hirai, and A.~Matsumoto, ``Comparative study of
  robotics curricula,'' \emph{{IEEE} Transactions on Education}, vol.~64,
  no.~3, pp. 283--291, Aug. 2021. [Online]. Available:
  \url{https://doi.org/10.1109/te.2020.3041667}
\BIBentrySTDinterwordspacing

\bibitem{Papert1980}
S.~Papert, \emph{Mindstorms children, computers, and powerful ideas}.\hskip 1em
  plus 0.5em minus 0.4em\relax London, England: Basic Books, 1980.

\bibitem{Betz2022}
J.~Betz, H.~Zheng, A.~Liniger, U.~Rosolia, P.~Karle, M.~Behl, V.~Krovi, and
  R.~Mangharam, ``Autonomous vehicles on the edge: A survey on autonomous
  vehicle racing,'' 2022.

\bibitem{Medeiros2019}
\BIBentryALTinterwordspacing
R.~P. Medeiros, G.~L. Ramalho, and T.~P. Falcao, ``A systematic literature
  review on teaching and learning introductory programming in higher
  education,'' \emph{{IEEE} Transactions on Education}, vol.~62, no.~2, pp.
  77--90, May 2019. [Online]. Available:
  \url{https://doi.org/10.1109/te.2018.2864133}
\BIBentrySTDinterwordspacing

\bibitem{Burguillo2010}
\BIBentryALTinterwordspacing
J.~C. Burguillo, ``Using game theory and competition-based learning to
  stimulate student motivation and performance,'' \emph{Computers \&Education},
  vol.~55, no.~2, pp. 566--575, Sep. 2010. [Online]. Available:
  \url{https://doi.org/10.1016/j.compedu.2010.02.018}
\BIBentrySTDinterwordspacing

\bibitem{Shim2017}
\BIBentryALTinterwordspacing
J.~Shim, D.~Kwon, and W.~Lee, ``The effects of a robot game environment on
  computer programming education for elementary school students,'' \emph{{IEEE}
  Transactions on Education}, vol.~60, no.~2, pp. 164--172, May 2017. [Online].
  Available: \url{https://doi.org/10.1109/te.2016.2622227}
\BIBentrySTDinterwordspacing

\bibitem{Bers2014}
M.~U. Bers, L.~Flannery, E.~R. Kazakoff, and A.~Sullivan,
  ``\BIBforeignlanguage{en}{Computational thinking and tinkering: Exploration
  of an early childhood robotics curriculum},''
  \emph{\BIBforeignlanguage{en}{Comput. Educ.}}, vol.~72, pp. 145--157, Mar.
  2014.

\bibitem{Diago2021}
P.~D. Diago, J.~A. Gonz{\'a}lez-Calero, and D.~F. Y{\'a}{\~n}ez,
  ``\BIBforeignlanguage{en}{Exploring the development of mental rotation and
  computational skills in elementary students through educational robotics},''
  \emph{\BIBforeignlanguage{en}{Int. J. Child Comput. Interact.}}, no. 100388,
  p. 100388, Aug. 2021.

\bibitem{Kucuk2020}
S.~Kucuk and B.~Sisman, ``\BIBforeignlanguage{en}{Students' attitudes towards
  robotics and {STEM}: Differences based on gender and robotics experience},''
  \emph{\BIBforeignlanguage{en}{Int. J. Child Comput. Interact.}}, vol. 23-24,
  no. 100167, p. 100167, Jun. 2020.

\bibitem{Bakala2021}
E.~Bakala, A.~Gerosa, J.~P. Hourcade, and G.~Tejera,
  ``\BIBforeignlanguage{en}{Preschool children, robots, and computational
  thinking: A systematic review},'' \emph{\BIBforeignlanguage{en}{Int. J. Child
  Comput. Interact.}}, vol.~29, no. 100337, p. 100337, Sep. 2021.

\bibitem{Jung2018}
\BIBentryALTinterwordspacing
S.~E. Jung and E.-s. Won, ``Systematic review of research trends in robotics
  education for young children,'' \emph{Sustainability}, vol.~10, no.~4, 2018.
  [Online]. Available: \url{https://www.mdpi.com/2071-1050/10/4/905}
\BIBentrySTDinterwordspacing

\bibitem{Di_Lieto2017}
M.~C. Di~Lieto, E.~Inguaggiato, E.~Castro, F.~Cecchi, G.~Cioni, M.~Dell'Omo,
  C.~Laschi, C.~Pecini, G.~Santerini, G.~Sgandurra, and P.~Dario, ``Educational
  robotics intervention on executive functions in preschool children: A pilot
  study,'' \emph{Comput. Human Behav.}, vol.~71, pp. 16--23, Jun. 2017.

\bibitem{Savard2016}
A.~Savard and V.~Freiman, ``\BIBforeignlanguage{en}{Investigating complexity to
  assess student learning from a robotics-based task},''
  \emph{\BIBforeignlanguage{en}{Digit. Exp. Math. Educ.}}, vol.~2, no.~2, pp.
  93--114, Sep. 2016.

\bibitem{Irigoyen2013}
E.~Irigoyen, E.~Larzabal, and R.~Priego, ``\BIBforeignlanguage{en}{Low-cost
  platforms used in control education: An educational case study},''
  \emph{\BIBforeignlanguage{en}{IFAC proc. vol.}}, vol.~46, no.~17, pp.
  256--261, 2013.

\bibitem{Esposito2017}
J.~M. Esposito, ``The state of robotics education: Proposed goals for
  positively transforming robotics education at postsecondary institutions,''
  \emph{IEEE Robot. Autom. Mag.}, vol.~24, no.~3, pp. 157--164, Sep. 2017.

\bibitem{Cielniak2013}
G.~Cielniak, N.~Bellotto, and T.~Duckett, ``Integrating mobile robotics and
  vision with undergraduate computer science,'' \emph{IEEE trans. educ.},
  vol.~56, no.~1, pp. 48--53, Feb. 2013.

\bibitem{Frank2018}
D.~Frank, K.~Kolotka, A.~Phillips, M.~Schulz, C.~Rigney, A.~Drown, R.~Stricko,
  K.~Harper, and R.~Freuler, ``Developing and improving a multi-element
  first-year engineering cornerstone autonomous robotics design project,'' in
  \emph{2017 {ASEE} Annual Conference \& Exposition Proceedings}.\hskip 1em
  plus 0.5em minus 0.4em\relax ASEE Conferences, 2018.

\bibitem{Hildebrandt2022}
C.~Hildebrandt, M.~von Stein, T.~Woodlief, and E.~Sebastian, ``Preparing
  software engineers to develop robot systems.''\hskip 1em plus 0.5em minus
  0.4em\relax New York, NY, USA: ACM, 2022.

\bibitem{Tani2017}
\BIBentryALTinterwordspacing
J.~Tani, L.~Paull, M.~T. Zuber, D.~Rus, J.~How, J.~Leonard, and A.~Censi,
  ``Duckietown: An innovative way to teach autonomy,'' in \emph{Educational
  Robotics in the Makers Era}.\hskip 1em plus 0.5em minus 0.4em\relax Springer
  International Publishing, 2017, pp. 104--121. [Online]. Available:
  \url{https://doi.org/10.1007/978-3-319-55553-9_8}
\BIBentrySTDinterwordspacing

\bibitem{Feng2018}
\BIBentryALTinterwordspacing
W.~Feng, Y.~Pei, B.~Hai, W.~Huang, X.~Gong, and F.~Zhu, ``Autonomous {RC}-car
  for education purpose in {iSTEM} projects,'' in \emph{2018 {IEEE} Intelligent
  Vehicles Symposium ({IV})}.\hskip 1em plus 0.5em minus 0.4em\relax {IEEE},
  Jun. 2018. [Online]. Available:
  \url{https://doi.org/10.1109/ivs.2018.8500633}
\BIBentrySTDinterwordspacing

\bibitem{Balaji2020}
\BIBentryALTinterwordspacing
B.~Balaji, S.~Mallya, S.~Genc, S.~Gupta, L.~Dirac, V.~Khare, G.~Roy, T.~Sun,
  Y.~Tao, B.~Townsend, E.~Calleja, S.~Muralidhara, and D.~Karuppasamy,
  ``{DeepRacer}: Autonomous racing platform for experimentation with sim2real
  reinforcement learning,'' in \emph{2020 {IEEE} International Conference on
  Robotics and Automation ({ICRA})}.\hskip 1em plus 0.5em minus 0.4em\relax
  {IEEE}, May 2020. [Online]. Available:
  \url{https://doi.org/10.1109/icra40945.2020.9197465}
\BIBentrySTDinterwordspacing

\bibitem{Karaman2017}
\BIBentryALTinterwordspacing
S.~Karaman, A.~Anders, M.~Boulet, J.~Connor, K.~Gregson, W.~Guerra, O.~Guldner,
  M.~Mohamoud, B.~Plancher, R.~Shin, and J.~Vivilecchia, ``Project-based,
  collaborative, algorithmic robotics for high school students: Programming
  self-driving race cars at {MIT},'' in \emph{2017 {IEEE} Integrated {STEM}
  Education Conference ({ISEC})}.\hskip 1em plus 0.5em minus 0.4em\relax
  {IEEE}, 2017. [Online]. Available:
  \url{https://doi.org/10.1109/isecon.2017.7910242}
\BIBentrySTDinterwordspacing

\bibitem{srinivasa2019}
S.~S. Srinivasa, P.~Lancaster, J.~Michalove, M.~Schmittle, C.~Summers,
  M.~Rockett, J.~R. Smith, S.~Choudhury, C.~Mavrogiannis, and F.~Sadeghi,
  ``Mushr: A low-cost, open-source robotic racecar for education and
  research,'' 2019.

\bibitem{Hart2014}
\BIBentryALTinterwordspacing
K.~Hart, C.~Montella, G.~Petitpas, D.~Schweisinger, A.~Shariati, B.~Sourbeer,
  T.~Trephan, and J.~Spletzer, ``{RoSCAR},'' in \emph{Proceedings of the 2014
  workshop on Mobile augmented reality and robotic technology-based systems -
  {MARS} {\textquotesingle}14}.\hskip 1em plus 0.5em minus 0.4em\relax {ACM}
  Press, 2014. [Online]. Available:
  \url{https://doi.org/10.1145/2609829.2609837}
\BIBentrySTDinterwordspacing

\bibitem{okelly2019}
M.~O'Kelly, V.~Sukhil, H.~Abbas, J.~Harkins, C.~Kao, Y.~V. Pant, R.~Mangharam,
  D.~Agarwal, M.~Behl, P.~Burgio, and M.~Bertogna, ``F1/10: An open-source
  autonomous cyber-physical platform,'' 2019.

\bibitem{Agnihotri2020}
\BIBentryALTinterwordspacing
A.~Agnihotri, M.~O{\textquotesingle}Kelly, R.~Mangharam, and H.~Abbas,
  ``Teaching autonomous systems at 1/10th-scale,'' in \emph{Proceedings of the
  51st {ACM} Technical Symposium on Computer Science Education}.\hskip 1em plus
  0.5em minus 0.4em\relax {ACM}, Feb. 2020, pp. 657--663. [Online]. Available:
  \url{https://doi.org/10.1145/3328778.3366796}
\BIBentrySTDinterwordspacing

\bibitem{Paul2009}
P.~Smith, ``Bridging the gap in engineering education through student
  competition,'' in \emph{2009 ICCAS-SICE}, 2009, pp. 1877--1877.

\bibitem{Hulls2020}
\BIBentryALTinterwordspacing
C.~C.~W. Hulls and C.~Rennick, ``Use of a cornerstone project to teach
  ill-structured software design in first year,'' \emph{{IEEE} Transactions on
  Education}, vol.~63, no.~2, pp. 98--107, May 2020. [Online]. Available:
  \url{https://doi.org/10.1109/te.2019.2959591}
\BIBentrySTDinterwordspacing

\bibitem{GomezdeGabriel2015}
\BIBentryALTinterwordspacing
J.~M.~G. de~Gabriel, A.~Mandow, J.~Fernandez-Lozano, and A.~Garcia-Cerezo,
  ``Mobile robot lab project to introduce engineering students to fault
  diagnosis in mechatronic systems,'' \emph{{IEEE} Transactions on Education},
  vol.~58, no.~3, pp. 187--193, Aug. 2015. [Online]. Available:
  \url{https://doi.org/10.1109/te.2014.2358551}
\BIBentrySTDinterwordspacing

\bibitem{Anwar2019}
\BIBentryALTinterwordspacing
S.~Anwar, N.~A. Bascou, M.~Menekse, and A.~Kardgar, ``A systematic review of
  studies on educational robotics,'' \emph{Journal of Pre-College Engineering
  Education Research (J-{PEER})}, vol.~9, no.~2, Jul. 2019. [Online].
  Available: \url{https://doi.org/10.7771/2157-9288.1223}
\BIBentrySTDinterwordspacing

\bibitem{Miller2008}
\BIBentryALTinterwordspacing
D.~P. Miller, I.~R. Nourbakhsh, and R.~Siegwart, ``Robots for education,'' in
  \emph{Springer Handbook of Robotics}.\hskip 1em plus 0.5em minus 0.4em\relax
  Springer Berlin Heidelberg, 2008, pp. 1283--1301. [Online]. Available:
  \url{https://doi.org/10.1007/978-3-540-30301-5_56}
\BIBentrySTDinterwordspacing

\bibitem{nvidiajetson}
\BIBentryALTinterwordspacing
{Connect Tech}, ``Jetson module comparison,'' Aug 2021. [Online]. Available:
  \url{https://connecttech.com/jetson/jetson-module-comparison/}
\BIBentrySTDinterwordspacing

\bibitem{Giang2019}
\BIBentryALTinterwordspacing
C.~Giang, A.~Piatti, and F.~Mondada, ``Heuristics for the development and
  evaluation of educational robotics systems,'' \emph{{IEEE} Transactions on
  Education}, vol.~62, no.~4, pp. 278--287, Nov. 2019. [Online]. Available:
  \url{https://doi.org/10.1109/te.2019.2912351}
\BIBentrySTDinterwordspacing

\bibitem{Levinson2011}
\BIBentryALTinterwordspacing
J.~Levinson, J.~Askeland, J.~Becker, J.~Dolson, D.~Held, S.~Kammel, J.~Z.
  Kolter, D.~Langer, O.~Pink, V.~Pratt, M.~Sokolsky, G.~Stanek, D.~Stavens,
  A.~Teichman, M.~Werling, and S.~Thrun, ``Towards fully autonomous driving:
  Systems and algorithms,'' in \emph{2011 {IEEE} Intelligent Vehicles Symposium
  ({IV})}.\hskip 1em plus 0.5em minus 0.4em\relax {IEEE}, Jun. 2011. [Online].
  Available: \url{https://doi.org/10.1109/ivs.2011.5940562}
\BIBentrySTDinterwordspacing

\bibitem{Redmon2016}
J.~Redmon, S.~Divvala, R.~Girshick, and A.~Farhadi, ``You only look once:
  Unified, real-time object detection,'' in \emph{2016 IEEE Conference on
  Computer Vision and Pattern Recognition (CVPR)}, 2016, pp. 779--788.

\bibitem{sezer2012novel}
V.~Sezer and M.~Gokasan, ``A novel obstacle avoidance algorithm:“follow the
  gap method”,'' \emph{Robotics and Autonomous Systems}, vol.~60, no.~9, pp.
  1123--1134, 2012.

\bibitem{Heilmeier2019}
\BIBentryALTinterwordspacing
A.~Heilmeier, A.~Wischnewski, L.~Hermansdorfer, J.~Betz, M.~Lienkamp, and
  B.~Lohmann, ``Minimum curvature trajectory planning and control for an
  autonomous race car,'' \emph{Vehicle System Dynamics}, vol.~58, no.~10, pp.
  1497--1527, Jun. 2019. [Online]. Available:
  \url{https://doi.org/10.1080/00423114.2019.1631455}
\BIBentrySTDinterwordspacing

\bibitem{Werling2010}
\BIBentryALTinterwordspacing
M.~Werling, J.~Ziegler, S.~Kammel, and S.~Thrun, ``Optimal trajectory
  generation for dynamic street scenarios in a frenet frame,'' in \emph{2010
  {IEEE} International Conference on Robotics and Automation}.\hskip 1em plus
  0.5em minus 0.4em\relax {IEEE}, May 2010. [Online]. Available:
  \url{https://doi.org/10.1109/robot.2010.5509799}
\BIBentrySTDinterwordspacing

\bibitem{Stahl2019}
\BIBentryALTinterwordspacing
T.~Stahl, A.~Wischnewski, J.~Betz, and M.~Lienkamp, ``Multilayer graph-based
  trajectory planning for race vehicles in dynamic scenarios,'' in \emph{2019
  {IEEE} Intelligent Transportation Systems Conference ({ITSC})}.\hskip 1em
  plus 0.5em minus 0.4em\relax {IEEE}, Oct. 2019. [Online]. Available:
  \url{https://doi.org/10.1109/itsc.2019.8917032}
\BIBentrySTDinterwordspacing

\bibitem{coulter1992implementation}
R.~C. Coulter, ``Implementation of the pure pursuit path tracking algorithm,''
  Carnegie-Mellon UNIV Pittsburgh PA Robotics INST, Tech. Rep., 1992.

\bibitem{thrun2006stanley}
S.~Thrun, M.~Montemerlo, H.~Dahlkamp, D.~Stavens, A.~Aron, J.~Diebel, P.~Fong,
  J.~Gale, M.~Halpenny, G.~Hoffmann \emph{et~al.}, ``Stanley: The robot that
  won the darpa grand challenge,'' \emph{Journal of field Robotics}, vol.~23,
  no.~9, pp. 661--692, 2006.

\bibitem{Katriniok2013}
\BIBentryALTinterwordspacing
A.~Katriniok, J.~P. Maschuw, F.~Christen, L.~Eckstein, and D.~Abel, ``Optimal
  vehicle dynamics control for combined longitudinal and lateral autonomous
  vehicle guidance,'' in \emph{2013 European Control Conference ({ECC})}.\hskip
  1em plus 0.5em minus 0.4em\relax {IEEE}, Jul. 2013. [Online]. Available:
  \url{https://doi.org/10.23919/ecc.2013.6669331}
\BIBentrySTDinterwordspacing

\bibitem{Kato2015}
\BIBentryALTinterwordspacing
S.~Kato, E.~Takeuchi, Y.~Ishiguro, Y.~Ninomiya, K.~Takeda, and T.~Hamada, ``An
  open approach to autonomous vehicles,'' \emph{{IEEE} Micro}, vol.~35, no.~6,
  pp. 60--68, Nov. 2015. [Online]. Available:
  \url{https://doi.org/10.1109/mm.2015.133}
\BIBentrySTDinterwordspacing

\bibitem{Yang2021}
\BIBentryALTinterwordspacing
G.~Yang, Y.~Xue, L.~Meng, P.~Wang, Y.~Shi, Q.~Yang, and Q.~Dong, ``Survey on
  autonomous vehicle simulation platforms,'' in \emph{2021 8th International
  Conference on Dependable Systems and Their Applications ({DSA})}.\hskip 1em
  plus 0.5em minus 0.4em\relax {IEEE}, Aug. 2021. [Online]. Available:
  \url{https://doi.org/10.1109/dsa52907.2021.00100}
\BIBentrySTDinterwordspacing

\bibitem{Althoff2017}
\BIBentryALTinterwordspacing
M.~Althoff, M.~Koschi, and S.~Manzinger, ``{CommonRoad}: Composable benchmarks
  for motion planning on roads,'' in \emph{2017 {IEEE} Intelligent Vehicles
  Symposium ({IV})}.\hskip 1em plus 0.5em minus 0.4em\relax {IEEE}, Jun. 2017.
  [Online]. Available: \url{https://doi.org/10.1109/ivs.2017.7995802}
\BIBentrySTDinterwordspacing

\bibitem{brockman2016openai}
G.~Brockman, V.~Cheung, L.~Pettersson, J.~Schneider, J.~Schulman, J.~Tang, and
  W.~Zaremba, ``Openai gym,'' \emph{arXiv preprint arXiv:1606.01540}, 2016.

\bibitem{Guodong2020}
G.~Rong, B.~H. Shin, H.~Tabatabaee, Q.~Lu, S.~Lemke, M.~Mozeiko, E.~Boise,
  G.~Uhm, M.~Gerow, S.~Mehta, E.~Agafonov, T.~H. Kim, E.~Sterner, K.~Ushiroda,
  M.~Reyes, D.~Zelenkovsky, and S.~Kim, ``{LGSVL} simulator: A high fidelity
  simulator for autonomous driving,'' in \emph{2020 {IEEE} 23rd International
  Conference on Intelligent Transportation Systems ({ITSC})}.\hskip 1em plus
  0.5em minus 0.4em\relax {IEEE}, Sep. 2020.

\bibitem{Bulsara2020}
\BIBentryALTinterwordspacing
A.~Bulsara, A.~Raman, S.~Kamarajugadda, M.~Schmid, and V.~N. Krovi, ``Obstacle
  avoidance using model predictive control: An implementation and validation
  study using scaled vehicles,'' in \emph{{SAE} Technical Paper Series}.\hskip
  1em plus 0.5em minus 0.4em\relax {SAE} International, Apr. 2020. [Online].
  Available: \url{https://doi.org/10.4271/2020-01-0109}
\BIBentrySTDinterwordspacing

\bibitem{Ivanov2020}
\BIBentryALTinterwordspacing
R.~Ivanov, T.~J. Carpenter, J.~Weimer, R.~Alur, G.~J. Pappas, and I.~Lee,
  ``Case study,'' in \emph{Proceedings of the 23rd International Conference on
  Hybrid Systems: Computation and Control}.\hskip 1em plus 0.5em minus
  0.4em\relax {ACM}, Apr. 2020. [Online]. Available:
  \url{https://doi.org/10.1145/3365365.3382216}
\BIBentrySTDinterwordspacing

\bibitem{jain2020}
A.~Jain, M.~O'Kelly, P.~Chaudhari, and M.~Morari, ``Bayesrace: Learning to race
  autonomously using prior experience,'' 2020.

\bibitem{Tatulea2020}
\BIBentryALTinterwordspacing
A.~T{\u{a}}tulea-Codrean, T.~Mariani, and S.~Engell, ``Design and simulation of
  a machine-learning and model predictive control approach to autonomous race
  driving for the f1/10 platform,'' \emph{{IFAC}-{PapersOnLine}}, vol.~53,
  no.~2, pp. 6031--6036, 2020. [Online]. Available:
  \url{https://doi.org/10.1016/j.ifacol.2020.12.1669}
\BIBentrySTDinterwordspacing

\bibitem{Rosolia2020}
\BIBentryALTinterwordspacing
U.~Rosolia and F.~Borrelli, ``Learning how to autonomously race a car: A
  predictive control approach,'' \emph{{IEEE} Transactions on Control Systems
  Technology}, vol.~28, no.~6, pp. 2713--2719, Nov. 2020. [Online]. Available:
  \url{https://doi.org/10.1109/tcst.2019.2948135}
\BIBentrySTDinterwordspacing

\bibitem{Pagot2020}
\BIBentryALTinterwordspacing
E.~Pagot, M.~Piccinini, and F.~Biral, ``Real-time optimal control of an
  autonomous {RC} car with minimum-time maneuvers and a novel kineto-dynamical
  model,'' in \emph{2020 {IEEE}/{RSJ} International Conference on Intelligent
  Robots and Systems ({IROS})}.\hskip 1em plus 0.5em minus 0.4em\relax {IEEE},
  Oct. 2020. [Online]. Available:
  \url{https://doi.org/10.1109/iros45743.2020.9340640}
\BIBentrySTDinterwordspacing

\bibitem{Brunnbauer2019}
\BIBentryALTinterwordspacing
A.~Brunnbauer and M.~Bader, ``Traffic cone based self-localization on a 1:10
  race car,'' 2019. [Online]. Available:
  \url{https://openlib.tugraz.at/download.php?id=5d09dba11efec&location=medra}
\BIBentrySTDinterwordspacing

\bibitem{Gotlib2019}
\BIBentryALTinterwordspacing
A.~Gotlib, K.~Lukojc, and M.~Szczygielski, ``Localization-based software
  architecture for 1:10 scale autonomous car,'' in \emph{2019 International
  Interdisciplinary {PhD} Workshop ({IIPhDW})}.\hskip 1em plus 0.5em minus
  0.4em\relax {IEEE}, May 2019. [Online]. Available:
  \url{https://doi.org/10.1109/iiphdw.2019.8755418}
\BIBentrySTDinterwordspacing

\bibitem{xiao2021ikd}
X.~Xiao, J.~Biswas, and P.~Stone, ``Learning inverse kinodynamics for accurate
  high-speed off-road navigation on unstructured terrain,'' \emph{IEEE Robotics
  and Automation Letters (RA-L)}, pp. 1--7, 2021.

\bibitem{karnan2022viikd}
\BIBentryALTinterwordspacing
H.~Karnan, K.~S. Sikand, P.~Atreya, S.~Rabiee, X.~Xiao, G.~Warnell, P.~Stone,
  and J.~Biswas, ``{ VI-IKD: High-Speed Accurate Off-Road Navigation using
  Learned Visual-Inertial Inverse Kinodynamics },'' arXiv Preprint
  arXiv:2203.15983, 2022. [Online]. Available:
  \url{https://arxiv.org/abs/2203.15983}
\BIBentrySTDinterwordspacing

\bibitem{wei2021onevision}
J.~Wei, T.~Li, S.~Chaudhuri, I.~Dillig, and J.~Biswas, ``Onevision: Centralized
  to distributed controller synthesis with delay compensation,'' in
  \emph{Intelligent Robots and Systems (IROS), IEEE/RSJ International
  Conference on}, 2021, pp. 398--405.

\end{thebibliography}
